\documentclass[10pt,twocolumn,letterpaper]{article}

\pdfoutput=1

\usepackage{iccv}
\usepackage{times}
\usepackage{epsfig}
\usepackage{graphicx}
\usepackage{amsmath}
\usepackage{amssymb}
\usepackage[usenames,dvipsnames]{xcolor}
\usepackage{csvsimple}
\usepackage{comment}
\usepackage{soul}
\usepackage{multirow}

\usepackage[boldmath]{numprint}
\usepackage{array, siunitx, booktabs}
\usepackage{enumitem}
\setlist[itemize]{parsep=0pt,topsep=2pt,itemsep=2pt}
\setlist[enumerate]{parsep=0pt,topsep=1pt,itemsep=1pt}

\usepackage{booktabs}
\usepackage[]{pgfplots,pgfplotstable}
\usepackage[accsupp]{axessibility}  

\usepackage[]{tikz}
\pgfplotsset{compat=1.10}
\pgfplotsset{every axis label/.append style={font=\large}}
\pgfplotsset{every tick label/.append style={font=\large}}
\usetikzlibrary{intersections}
\usepgfplotslibrary{fillbetween}

\pgfplotstableset{
    create on use/Ymd15/.style={create col/copy column from table={data/s3dis/s3dis_mc_dropout_init_set_15_closest_dino.txt}{y}}
}

\pgfplotstableset{
    create on use/Yredal15/.style={create col/copy column from table={data/s3dis/s3dis_ReDAL_init_set_15_closest_dino.txt}{y}}
}

\pgfplotstableset{
    create on use/Yrand15/.style={create col/copy column from table={data/s3dis/s3dis_random_init_set_15_closest_dino.txt}{y}}
}

\pgfplotstableset{
    create on use/Ysege15/.style={create col/copy column from table={data/s3dis/s3dis_segment_entropy_init_set_15_closest_dino.txt}{y}}
}

\pgfplotstableset{
    create on use/Ysofc15/.style={create col/copy column from table={data/s3dis/s3dis_softmax_confidence_init_set_15_closest_dino.txt}{y}}
}

\pgfplotstableset{
    create on use/Ycs10/.style={create col/copy column from table={data/s3dis/s3dis_coreset_init_set_10_closest_dino.txt}{y}}
}

\pgfplotstableset{
    create on use/Yredal10/.style={create col/copy column from table={data/s3dis/s3dis_ReDAL_init_set_10_closest_dino.txt}{y}}
}

\pgfplotstableset{
    create on use/Ysege10/.style={create col/copy column from table={data/s3dis/s3dis_segment_entropy_init_set_10_closest_dino.txt}{y}}
}

\pgfplotstableset{
    create on use/Ysofc10/.style={create col/copy column from table={data/s3dis/s3dis_softmax_confidence_init_set_10_closest_dino.txt}{y}}
}

\pgfplotstableset{
    create on use/Ysofe10/.style={create col/copy column from table={data/s3dis/s3dis_softmax_entropy_init_set_10_closest_dino.txt}{y}}
}

\pgfplotstableset{
    create on use/Ysofm10/.style={create col/copy column from table={data/s3dis/s3dis_softmax_margin_init_set_10_closest_dino.txt}{y}}
}

\pgfplotstableset{
    create on use/Yredal/.style={create col/copy column from table={data/s3dis/s3dis_ReDAL_init_set_6_closest_dino.txt}{y}}
}
\pgfplotstableset{
    create on use/Yrandom/.style={create col/copy column from table={data/s3dis/s3dis_random_init_set_6_closest_dino.txt}{y}}
}
\pgfplotstableset{
    create on use/Ycs/.style={create col/copy column from table={data/s3dis/s3dis_coreset_init_set_6_closest_dino.txt}{y}}
}
\pgfplotstableset{
    create on use/Ymc/.style={create col/copy column from table={data/s3dis/s3dis_mc_dropout_init_set_6_closest_dino.txt}{y}}
}
\pgfplotstableset{
    create on use/Yse/.style={create col/copy column from table={data/s3dis/s3dis_segment_entropy_init_set_6_closest_dino.txt}{y}}
}
\pgfplotstableset{
    create on use/Ysoc/.style={create col/copy column from table={data/s3dis/s3dis_softmax_confidence_init_set_6_closest_dino.txt}{y}}
}
\pgfplotstableset{
    create on use/Ysoe/.style={create col/copy column from table={data/s3dis/s3dis_softmax_entropy_init_set_6_closest_dino.txt}{y}}
}
\pgfplotstableset{
    create on use/Ysom/.style={create col/copy column from table={data/s3dis/s3dis_softmax_margin_init_set_6_closest_dino.txt}{y}}
}

\usepackage{pgfplots}
\usepackage[nomessages]{fp}
\usepackage{pgfplots}
\usepackage{pgfkeys}
\newenvironment{customlegend}[1][]{%
    \begingroup
    \csname pgfplots@init@cleared@structures\endcsname
    \pgfplotsset{#1}%
}{%
    \csname pgfplots@createlegend\endcsname
    \endgroup
}%
\def\addlegendimage{\csname pgfplots@addlegendimage\endcsname}

\usepackage{subcaption}
\usepackage{color}

\newif\ifshowedits

\newcommand{\addeditor}[3]{%
  \definecolor{#1color}{rgb}{#3}
  \expandafter\newcommand\csname #1\endcsname[1]{%
  \ifshowedits
    {\color{#1color} ##1}%
  \else{##1}\fi
  }%
  \expandafter\newcommand\csname #1rmk\endcsname[1]{%
\ifshowedits
    {\color{#1color} {\bf [#2: ##1]}}
\fi}%
  \expandafter\newcommand\csname #1rpl\endcsname[2]{%
  \ifshowedits
    {\color{#1color} ##1 \sout{##2}}
  \else{##1}\fi
  }%
}

\newcommand{\mycomment}[1]{}

\DeclareMathOperator*{\argmin}{arg\,min}
\DeclareMathOperator*{\argmax}{arg\,max}

\DeclareFontFamily{U}{mathx}{\hyphenchar\font45}
\DeclareFontShape{U}{mathx}{m}{n}{
      <5> <6> <7> <8> <9> <10>
      <10.95> <12> <14.4> <17.28> <20.74> <24.88>
      mathx10
      }{}
\DeclareSymbolFont{mathx}{U}{mathx}{m}{n}
\DeclareFontSubstitution{U}{mathx}{m}{n}
\DeclareMathAccent{\widebar}{0}{mathx}{"73}

\newcommand{\dino}[2]{\psi_{#1}^{#2}}

\newcommand{\ssp}[1]{\,{#1}\,}

\addeditor{vincent}{VL}{0.0, 0.5, 0.0}
\addeditor{renaud}{RM}{1.0, 0.5, 0}
\addeditor{nermin}{NS}{0.5, 0.0, 0.5}
\newcommand{\OS}[1]{\textcolor{NavyBlue}{#1}}
\newcommand{\OSc}[1]{\textbf{\textcolor{NavyBlue}{[OS:#1]}}}
\newcommand{\gilles}[1]{\textcolor{teal}{#1}}
\newcommand{\gillesc}[1]{\textcolor{teal}{\bf[Gilles: #1]}}

\showeditsfalse
\renewcommand{\OS}[1]{#1}
\renewcommand{\OSc}[1]{}
\renewcommand{\gilles}[1]{#1}
\renewcommand{\gillesc}[1]{}

\def\ours{{SeedAL}\xspace} 

\def\random{rand\xspace}
\def\mcdropout{MC-Drop\xspace}
\def\softconf{S-conf\xspace}
\def\softmargin{S-margin\xspace}
\def\softentropy{S-ent\xspace}
\def\coreset{CoreSet\xspace}
\def\segent{SegEnt\xspace}
\def\redal{ReDAL\xspace}
\def\KMfirst{KMcentroid\xspace}
\def\KMsecond{KMfurthest\xspace}

\def\depthcontrast{{DepthContrast}\xspace} 
\def\also{{ALSO}\xspace} 
\def\segcontrast{{SegContrast}\xspace} 
\def\moco{{MoCo-v3}\xspace}
\def\dino{{DINO}\xspace}

\renewcommand{\paragraph}[1]{\medskip\noindent\textbf{#1}~}

\usepackage[pagebackref=true,breaklinks=true,letterpaper=true,colorlinks,bookmarks=false]{hyperref}

\iccvfinalcopy 

\def\iccvPaperID{1856} 

\ificcvfinal\fi

\begin{document}

\title{You Never Get a Second Chance To Make a Good First Impression: \\ Seeding Active Learning for 3D Semantic Segmentation}

\author{Nermin Samet$^1$ \qquad
Oriane Siméoni$^2$ \qquad
Gilles Puy$^2$ \qquad
Georgy Ponimatkin$^1$ \\ 
Renaud Marlet$^{1,2}$ \qquad
Vincent Lepetit$^1$ \\
\hspace{-5mm}\textsuperscript{1}LIGM, Ecole des Ponts, Univ Gustave Eiffel, CNRS, Marne-la-Vall\'ee, France
\hspace{2.5mm}\textsuperscript{2}Valeo.ai, Paris, France}

\author{\hspace{-3mm}Nermin Samet$^1$,
Oriane Siméoni$^2$,
Gilles Puy$^2$,
Georgy Ponimatkin$^1$, 
Renaud Marlet$^{1,2}$,
Vincent Lepetit$^1$ \\[2mm]
\hspace{-5mm}\textsuperscript{1}LIGM, Ecole des Ponts, Univ Gustave Eiffel, CNRS, Marne-la-Vall\'ee, France
\hspace{2.5mm}\textsuperscript{2}Valeo.ai, Paris, France
}

\maketitle
\ificcvfinal\thispagestyle{empty}\fi

\begin{abstract}
  
We propose \ours, a method to seed active learning for efficient annotation of 3D point clouds for semantic segmentation. Active Learning~(AL) iteratively selects relevant data fractions to annotate within a given budget, but requires a first fraction of the dataset~(a 'seed') to be already annotated to estimate the benefit of annotating other data fractions. We first show that the choice of the seed can significantly affect the performance of many AL methods. 
We then propose a method for automatically constructing a seed that will ensure good performance for AL. Assuming that images of the point clouds are available, which is common, our method relies on powerful unsupervised image features to measure the diversity of the point clouds. It selects the point clouds for the seed by optimizing the diversity  under an annotation budget, which can be done by solving a linear optimization problem. Our experiments demonstrate the effectiveness of our approach compared to random seeding and existing methods on both the S3DIS and SemanticKitti datasets. Code is available at \url{https://github.com/nerminsamet/seedal}.

\end{abstract}

\section{Introduction}
\label{sec:introduction}

\begin{figure}
    \begin{subfigure}{.5\linewidth}
      \centering
        \centering
\begin{tikzpicture}
    \tikzstyle{every node}=[font=\small]
    \begin{axis}[
        width=4.7cm,
        height=4.2cm,
        font=\footnotesize,
        xtick={3,5,7,9},
        ylabel=mIoU (\%),
        label style={font=\footnotesize},
        tick label style={font=\footnotesize},
        title style={yshift=-1.ex,},
        title=Core-Set,
        legend pos=south east,
        grid=major,
        legend style={nodes={scale=0.7, transform shape}},
        legend cell align={left}
    ]
    \foreach \file in {1,2,...,14,16,17,18,19,20}
    \addplot [color=RoyalBlue, mark=+, dashed, forget plot] table [y=y,x=x]{data/s3dis/s3dis_coreset_init_set_\file.txt};    
    \addplot [color=RoyalBlue, mark=+, dashed] table [y=y,x=x]{data/s3dis/s3dis_coreset_init_set_1.txt};
    \addlegendentry{random}; 
    \addplot [color=RoyalBlue, ultra thick, mark=+] table [y=mean,x=x]{data/s3dis/s3dis_coreset_init_set_1-20-mean-std.txt};
    \addlegendentry{random avg.\!\!};

    \addplot [color=purple, mark=+, ultra thick] table [y=y,x=x]{data/s3dis/s3dis_coreset_our_full_method.txt};
    \addlegendentry{\ours};
    \end{axis}
\end{tikzpicture}
      \label{fig:coreset-rand}
    \end{subfigure}%
    \begin{subfigure}{.5\linewidth}
      \centering
        \centering
\begin{tikzpicture}
    \tikzstyle{every node}=[font=\small]
    \begin{axis}[
        ymax=53,
        width=4.7cm,
        height=4.2cm,
        font=\footnotesize,
        xtick={3,5,7,9},
        label style={font=\footnotesize},
        tick label style={font=\footnotesize},
        title style={yshift=-1.ex,},
        title=ReDAL,
        legend pos=south east,
        grid=major,
        legend style={nodes={scale=0.7, transform shape}},
        legend cell align={left}
    ]
    \foreach \file in {1,2,...,14,16,17,18,19,20}
    \addplot [color=RoyalBlue, mark=+, dashed, forget plot] table [y=y,x=x]{data/s3dis/s3dis_ReDAL_init_set_\file.txt};
    \addplot [color=RoyalBlue, mark=+, dashed] table [y=y,x=x]{data/s3dis/s3dis_ReDAL_init_set_1.txt};
    \addlegendentry{random}; 
    
    \addplot [color=RoyalBlue, mark=+, ultra thick] table [y=mean,x=x]{data/s3dis/s3dis_ReDAL_init_set_1-20-mean-std.txt};
    \addlegendentry{random avg.\!\!};

    \addplot [color=purple, mark=+, ultra thick] table [y=y,x=x]{data/s3dis/s3dis_ReDAL_our_full_method.txt};
    \addlegendentry{\ours}; 
    \end{axis}
\end{tikzpicture}
      \label{fig:redal-rand}
    \end{subfigure}
    \\[-25pt]
    
    \begin{subfigure}{.5\linewidth}
      \centering
      \centering
\begin{tikzpicture}
    \tikzstyle{every node}=[font=\small]
    \begin{axis}[
        width=4.7cm,
        height=4.2cm,
        font=\footnotesize,
        xtick={3,5,7,9},
        xlabel=\% of labeled points,
        ylabel=mIoU (\%),
        label style={font=\footnotesize},
        tick label style={font=\footnotesize},
        title style={yshift=-1.ex,},
        title=MC Dropout,
        legend pos=south east,
        grid=major,
        legend style={nodes={scale=0.7, transform shape}},
        legend cell align={left}
    ]
    \foreach \file in {1,2,...,14,16,17,18,19,20}
    \addplot [color=RoyalBlue, mark=+,dashed,forget plot] table [y=y,x=x]{data/s3dis/s3dis_mc_dropout_init_set_\file.txt};    
    \addplot [color=RoyalBlue, mark=+,dashed] table [y=y,x=x]{data/s3dis/s3dis_mc_dropout_init_set_1.txt};
    \addlegendentry{random};  
    
    \addplot [color=RoyalBlue, mark=+, ultra thick] table [y=mean,x=x]{data/s3dis/s3dis_mc_dropout_init_set_1-20-mean-std.txt};
    \addlegendentry{random avg.\!\!};

    \addplot [color=purple, mark=+, ultra thick] table [y=y,x=x]{data/s3dis/s3dis_mc_dropout_our_full_method.txt};
    \addlegendentry{\ours};
    \end{axis}
\end{tikzpicture}
    \end{subfigure}%
    \begin{subfigure}{.5\linewidth}
      \centering
        \centering
\begin{tikzpicture}
    \tikzstyle{every node}=[font=\small]
    \begin{axis}[
        width=4.7cm,
        height=4.2cm,
        font=\footnotesize,
        xtick={3,5,7,9},
        xlabel=$\%$ of labeled points,
        label style={font=\footnotesize},
        tick label style={font=\footnotesize},
        title style={yshift=-1.ex,},
        title=Segment Entropy,
        grid=major,
        legend style={nodes={scale=0.7, transform shape}},
        legend pos=south east,
        legend cell align={left},
    ]
    \foreach \file in {1,2,...,14,16,17,18,19,20}
    \addplot [color=RoyalBlue, mark=+, dashed, forget plot] table [y=y,x=x]{data/s3dis/s3dis_segment_entropy_init_set_\file.txt};    
    \addplot [color=RoyalBlue, mark=+, dashed] table [y=y,x=x]{data/s3dis/s3dis_segment_entropy_init_set_1.txt};
    \addlegendentry{random}; 

    \addplot [color=RoyalBlue, mark=+, ultra thick] table [y=mean,x=x]{data/s3dis/s3dis_segment_entropy_init_set_1-20-mean-std.txt};
    \addlegendentry{random avg.\!\!};

    \addplot [color=purple, mark=+, ultra thick] table [y=y,x=x]{data/s3dis/s3dis_segment_entropy_our_full_method.txt};
    \addlegendentry{\ours};
    \end{axis}
\end{tikzpicture}
    \end{subfigure}
    \vspace{-30pt}
    \caption{\emph{Impact of active learning seed on performance.} We show the variability of results obtained with 20 different random seeds (blue dashed lines), within an initial annotation budget of 3\% of the dataset, when using various active learning methods
    for 3D semantic segmentation of S3DIS. We compare it to the result obtained with our seed selection strategy (solid red line), named \ours, which performs better or on par with the best (lucky) random seeds among 20, and ``protects'' from very bad (unlucky) random seeds. 
    }
    \label{fig:randomness}
\end{figure}
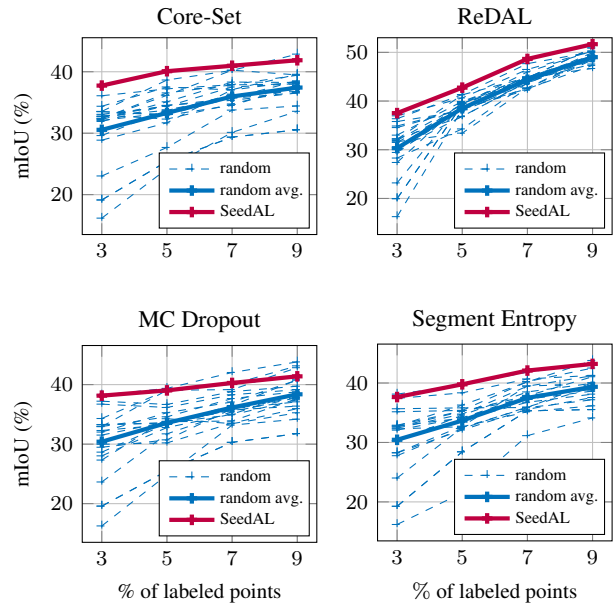

We are interested in the efficient annotation of sparse 3D point clouds (as captured indoors by depth cameras or outdoors by automotive lidars) for semantic segmentation.

Modern AI systems require training on large annotated datasets to reach a high performance on such a task. As annotating is costly (more than capturing the data itself), several approaches have been proposed to achieve a more frugal learning, such as semi-supervision \cite{engelen2020surveysemisup}, weak supervision \cite{zhou2017introweaksup}, few-shot \cite{parnami2022fewshot} and zero-shot learning~\cite{radford2021clip}, self-supervision \cite{longlong2021surveySSL, albelwi2022surveySSL} and, as studied here, active learning \cite{ren2021surveyAL}. 

\textbf{Active learning} (AL) methods iteratively select relevant fractions of a dataset to be annotated within a given budget so that, after a few iterations, the model learned on the annotated fractions reaches a performance close to the performance of the model learned on the fully-annotated dataset, although at a much lower annotation cost.  This AL selection typically targets the most uncertain \cite{joshi2009multiclass} or most diverse~\cite{coreset} data, that are assumed to have the most positive impact when annotated and used for training.

The criteria used in AL methods for selecting data to annotate generally require a preliminary fraction of the dataset to be already annotated. Assuming this fraction is representative enough of the rest of the dataset, it can be used to estimate the benefit of annotating other data fractions. But such a preliminary fraction is not available at the first AL iteration, when confronted to a completely unannotated dataset. There is thus a cold start problem, which is to determine a first fraction of the dataset to annotate --- the seed.

\textbf{Initial set.} Nevertheless, most AL publications pay little attention to the choice of this seed before iterating their AL method. They just pick a random fraction of the dataset. Although results are sometimes presented averaged over a few runs, it can be insufficient given the high variance level. As shown in \autoref{fig:randomness}, it is particularly true in our 3D semantic segmentation context. Besides, nothing prevents in practice from drawing an ``unlucky'' seed, leading to significant underperformance or annotation overheads.

To address this issue, our method, named \ours, automatically selects extremely good AL seeds, and thus also protects from drawing unlucky seeds, as illustrated on \autoref{fig:randomness}. Only few approaches have recently been proposed to construct such good seeds to get AL on track right away \cite{yuan2020cold, pourahmadi2021simple, lang2022bestpractice, hacohen2022budget, mahmood2022lowbudget, yehuda2022coveringlens, chen2022making, nath2022warmstart}, and they mostly regard text or image classification. To the best of our knowledge, no such AL seeding approach exists for 3D point clouds. 

\textbf{Self-supervised features}, for text or images \cite{devlin2019bert,he2020moco,caron2021dino}, are the key to most of these approaches. It thus seems natural to use self-supervised 3D features in our case. But existing pre-trained 3D backbones are not as versatile as their 2D counterpart, and provide lower quality features (see Sect.\,\ref{sec:2Dfeatratherthan3Dfeat}).
Our approach to seed AL for 3D point cloud is to leverage high-quality pretrained features \cite{caron2021dino} for images  that are views of the scanned scenes.

As we rely on image features like \cite{pourahmadi2021simple,lang2022bestpractice,hacohen2022budget,yehuda2022coveringlens,chen2022making}, it seems natural as well to reuse these AL seeding methods for 3D scenes.  But a direct transposition to 3D is not sensible because, in our case, the initial budget for the seed is not given in terms of number of images but in the size of scenes (number of 3D points): selecting a scene just because it contains a single image of interest 
would be suboptimal. 

\textbf{Our approach} integrates scenes and views to make a better use of the initial annotation budget.
While 2D approaches use feature clustering to guide seed selection, we show it is suboptimal in our context (2D views of 3D scenes) and propose a better formulation based on binary linear programming. As our algorithm has a higher complexity than clustering, we also develop heuristics to scale to large datasets by first extracting a small-enough pool of good candidates to select from.

Last, we observe that usual pretrained features, generally created from object-centric datasets such as ImageNet, fail to convey the diversity of complex multi-object scenes, by giving too much importance to big or repeated objects. We address this issue by analyzing images at patch level.

\textbf{Contributions.}
To the best of our knowledge, we are the first to address AL seeding for 3D point clouds: 
\begin{itemize}
    \item We study the sensitivity of AL seeding for the 3D semantic segmentation of point clouds and the relevance of 2D images features to select 3D scenes.
    \item Leveraging these studies, we propose a general approach to seed any AL method for point clouds originating from scenes with available image views.
    \item We present experimental evidence of the effectiveness of our AL seeding method, which consistently improves over random seeds and 2D-inspired baselines, requiring less iterations (thus less annotations) and/or reaching higher segmentation performance.
\end{itemize}

\section{Related Work}
\label{sec:relatedwork}

We discuss in this section different approaches to reduce the annotation work on point clouds, then focus on
Active Learning methods, and finally existing works on seed construction for active learning--even though none of them consider point cloud annotations.

\paragraph{Towards less supervision for 3D point clouds.}
The best approaches for 3D semantic segmentation of dense indoor point clouds~\cite{pointnext,pointmixer,pointtransformer,thomas2019KPConv} or sparse outdoor point clouds~\cite{2dpass,pvkd,cylinder3d,spvnas} are trained under full supervision, i.e., they require manual annotations of all points in all scenes, which is a notoriously costly task~\cite{semantickitti,genova2021learning}. Several approaches are currently explored to mitigate these costs. For example, one can leverage self-supervision~\cite{pointcontrast,depthcontrast,strl,gcc3d,segcontrast,slidr} to pre-train a neural network using many unlabeled data with an annotation-free pretext task and then fine-tune this network using few annotations for the task of interest. 
One can also rely on domain adaptation techniques~\cite{completelabel,cosmix,sfuda,Xu_2021_ICCV,st3d} to exploit existing annotated datasets on a source domain and avoid annotating a new dataset for a target domain. Weak supervision avoids the burden of complete point cloud annotation by, e.g., annotating a few points~\cite{oneclick,10xfewerlabels,sspcnet,zhang_aaai_2021}, annotating few regions in a scene \cite{liu2022less,Wei_2020_CVPR}, or using scene level labels~\cite{wypr,Wei_2020_CVPR}. Alternatively, 
active learning techniques, which we describe next, 
allow for a smart selection of the data to be annotated.

\paragraph{Active Learning for 3D point clouds.}
To minimize annotation costs, active learning strategies \cite{Brust2019,Geifman2017DeepAL,coreset} aim to select the most relevant data to annotate for the considered model and task.
Most selection strategies rely on diversity criteria~\cite{Geifman2017DeepAL,coreset,Ash2020Deep} or on a measure of the uncertainty of the model~\cite{Brust2019,softmax}.
In computer vision, active learning has been mostly developed for
classification tasks of 2D images~\cite{settles_active_2009}. It has been adapted to the semantic segmentation of 3D point clouds combined with markov random fields~\cite{luo2018mrf} and more recently deep learning for object detection~\cite{Feng2019DeepAL,luo2023exploring} and semantic segmentation~\cite{seg_ent,wu2021redal,hu2022lidal}.
Diversity of selected scenes has been ensured by selecting core sets~\cite{coreset, wu2021redal} in the feature space--in that space each scene is represented by aggregating the features of its corresponding points. 
The model uncertainty can be measured over each point class-scores~\cite{softmax,seg_ent}--e.g., using the entropy or the margin between the two highest scores--and averaged per scene in order to select the most confusing to the model. 
In \cite{seg_ent}, the authors improve results by computing the uncertainty scores at the level of pre-computed segments. Requiring several forward passes, ensembling techniques \cite{beluch2018power, gal2017deep, mcdropout} have been used to evaluate the model's confidence for 3D scenes~\cite{Feng2019DeepAL,wu2021redal}. 
Moving away from scene-based AL, \cite{Shi2021LabelEfficientPC,wu2021redal,hu2022lidal} propose region-based strategies--which mix  difficulty and diversity--that allow them to further reduce cost by selecting only local set of points to be annotated.
In this work, instead of proposing a new AL strategy, we show that a smart selection of the first set of data with \ours--usually over-looked and randomly selected--can boost all methods drastically.

\paragraph{Cold start problem.} 
The need for an initially annotated fraction of the data to bootstrap an AL method is similar to what has been identified as a cold start problem in collaborative filtering and recommendation systems, where new users start off with an empty profile~\cite{waltz1995pointing}.

In active learning, the term ``cold start problem'' has in fact been used with two close but different meanings. On the one hand \cite{houlsby2014coldstart, konyushkova2017lal,yehuda2022active}, it refers to the strong bias induced on the first AL iterations by a too small initial annotated set \cite{dasgupta2011twofaces}. In particular, uncertainty-based methods are not effective when trained with too little data~\cite{hacohen22active}. Note that, in this setting, a seed has nevertheless to be given as input, although if it is possible to set its size automatically to limit this bias, by deciding when to stop annotating the first data fraction constituting the seed~\cite{gao2020consistency}. On the other hand \cite{yuan2020cold,chen2022making}, the cold start problem also refers to the lack of a priori information to select a relevant seed in the first place, which is what we are addressing here.

\paragraph{Seeding active learning.}
Despite the high variance of performance with random seed selection, it appears it is difficult for AL approaches to create good seeds~\cite{chen2022making}.  In fact, only few recent approaches propose to automatically construct AL seeds \cite{yuan2020cold, pourahmadi2021simple, lang2022bestpractice, hacohen2022budget, mahmood2022lowbudget, yehuda2022coveringlens, chen2022making, nath2022warmstart}.  To the best of our knowledge, none applies to 3D point clouds, and there is only one concerning semantic segmentation \cite{nath2022warmstart}.

All approaches, except \cite{nath2022warmstart}, only address text or image classification, which is a much coarser-grained task than semantic segmentation. The general idea of most approaches for image classification \cite{pourahmadi2021simple, lang2022bestpractice, hacohen2022budget, yehuda2022coveringlens, chen2022making} is to use a self-supervised pretrained network, clustering the image features to estimate diversity groups and picking images close to cluster centers to get representatives, possibly focusing on hard-to-contrast and diverse data \cite{chen2022making} by exploiting contrastive self-supervised features \cite{chen2020improved}.
Alternatively, \cite{mahmood2022lowbudget} creates seeds using a core-set approach based on the Wasserstein distance between feature distributions of candidate data for the seed and the unannotated dataset. While this form of feature clustering or distance minimization between distributions also makes sense for 3D data, semantic segmentation requires extracting much more detailed information from each single datum to account for complex scenes (whether indoors or outdoors) that capture sets of objects, as opposed to object-centered pictures targeted by image classification in other approaches.

The only AL seeding approach we know for 3D, and in fact for semantic segmentation too, applies to dense 3D medical images (CT-scans) \cite{nath2022warmstart}, which significantly differ from sparse 3D point clouds captured by depth cameras or lidar scanners. The seeding strategy in \cite{nath2022warmstart} is based on a heuristic pseudo-labeling defined by a hand-parameterized, rule-based segmentation. It consists in thresholding CT data within a typical window of values for abdominal soft tissues, then extracting largest connected components and selecting major organs of the abdomen as foreground. This pseudo-labeling is specific to that peculiar kind of medical images and cannot be directly transposed to point clouds, although heuristic hand-designed pre-segments have also been used on 3D point clouds for label-efficient semantic segmentation~\cite{liu2022less}.

\section{Preliminary Study}
\label{sec:preliminary}

In this section, after introducing our notations and formally describing the problem, we highlight the sensitivity of various AL methods to the seed and then motivate the use of DINO \cite{caron2021dino} features for the selection of the seed. We use the results of this preliminary study to construct our method, \ours, which is fully described in the next section.

\subsection{Problem setup and notations}

Our method works in the following generic setting. We assume that a scene $i$ is captured by a depth sensor, providing a point cloud $P^i$, and by one or multiple cameras, providing several views $V^i_1, \ldots, V^i_{N_i}$ of the scene. For example, RGB-D cameras capture conveniently both the depth and the views simultaneously. 
Such a multi-modal setting is common, whether indoors \cite{s3dis,dai2017scannet} or outdoors \cite{semantickitti,nuscenes}.

Our dataset is made of a set of point clouds, with their respective views, and our goal is to select a good seed $S$, i.e., the initial set of the point clouds which will be fully annotated for the first active learning cycle.
As mentioned earlier, our approach for selecting $S$ is based on a pre-trained self-supervised image backbone.
We denote this backbone $\phi(\cdot)$ and use it to extract $D$-dimensional $\ell_2$-normalized features from each view: $\phi(V^s_n) \in \mathbb{R}^D$.

\begin{figure}[ht!]
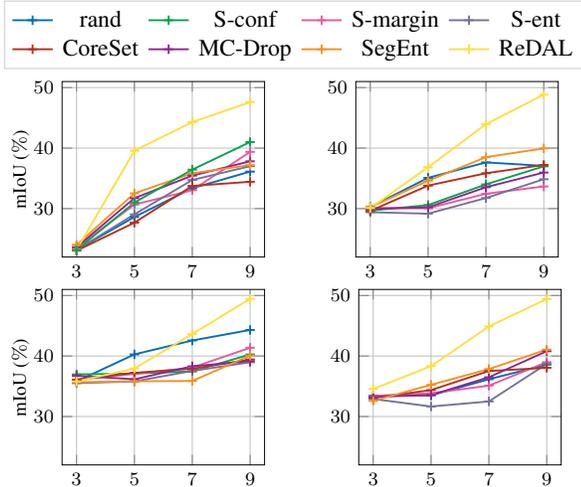

    \include{figs/tikz/random_seed_selection}
    \vspace{-10pt}
    \caption{\emph{Influence of the choice of the seed.} We visualize results obtained with several AL methods (introduced later in \autoref{sec:experiments}) when using 4 different random  seed. 
    With the exception of ReDAL, which is consistently better, the ranking of the methods varies with the seed.
    \vspace{-10pt}
    }
    \label{fig:random-seeds}
\end{figure}

\subsection{The high variance induced by random seeds}

Most AL publications use as AL seeds random fraction of the dataset --- which is one of the many issues for reproducibility~\cite{lang2022bestpractice,ji2023randomness}. A seed is typically picked within a maximum size, given as an initial annotation budget. 
For AL method evaluation, because of a high variance level due to the random AL seed, 
a common practice is 
to average results over  
few runs, randomizing the AL seed  
and the training iterations \cite{gao2020consistency, li2022noisestab, caramalau2021vit, lang2022bestpractice, pourahmadi2021simple, shui2020deep, ji2023randomness}. Even so, it can be not enough given the variance level. As shown in \autoref{fig:randomness}, it is particularly true in our problem.  
Besides, nothing prevents 
from drawing an ``unlucky'' seed, leading to significant underperformance or annotation overheads.

We first consider the common case of random seeds, i.e., picking random scenes until the annotation budget 
is reached. In \autoref{fig:randomness}, we illustrate the effect of drawing 20 such random seeds: it shows significant variations of performance, both at initial stage and after some iterations. 
Furthermore, as shown in \autoref{fig:random-seeds}, the performance hierarchy of scene-based AL methods differs depending on the seed.

A natural way to deal with both intra-method and inter-method variations is to consider the average and variance of the method performance over a large-enough number of random seeds. 
It allows evaluating the sensitivity of an AL method to the seed and it allows a better comparison of different AL methods, although still with a high variance.  However, it does not provide any practical hint how to choose and cold-start these methods. Besides, it is suboptimal as it  does not make the most of best seeds.

\begin{figure}
    \centering
\begin{tikzpicture}
    \tikzstyle{every node}=[font=\small]
    \begin{axis}[
        width=4.5cm,
        height=4.5cm,
        font=\footnotesize,
        xlabel=mIoU \emph{init} ($\%$),
        ylabel=mIoU \emph{closest} ($\%$),
        label style={font=\small},
        tick label style={font=\small},
        legend pos=north west,
        grid=major,
        legend style={nodes={scale=0.7, transform shape}},
        legend cell align={left}
    ]


    \addplot[color=black] coordinates {(35,35) (50,50)};
    \addplot [only marks, color=NavyBlue, mark=*] table [y=Ysoe,x=y]{data/s3dis/s3dis_softmax_entropy_init_set_6.txt};
    \addplot [only marks, color=NavyBlue, mark=*] table [y=Yredal,x=y]{data/s3dis/s3dis_ReDAL_init_set_6.txt};
    \addplot [only marks, color=NavyBlue, mark=*] table [y=Yrandom,x=y]{data/s3dis/s3dis_random_init_set_6.txt};
    \addplot [only marks, color=NavyBlue, mark=*] table [y=Ymc,x=y]{data/s3dis/s3dis_mc_dropout_init_set_6.txt};
    \addplot [only marks, color=NavyBlue, mark=*] table [y=Ycs,x=y]{data/s3dis/s3dis_coreset_init_set_6.txt};
    \addplot [only marks, color=NavyBlue, mark=*] table [y=Yse,x=y]{data/s3dis/s3dis_segment_entropy_init_set_6.txt};
    \addplot [only marks, color=NavyBlue, mark=*] table [y=Ysom,x=y]{data/s3dis/s3dis_softmax_margin_init_set_6.txt};
    
    
    \addplot [only marks, color=NavyBlue, mark=*] table [y=Ymd15,x=y]{data/s3dis/s3dis_mc_dropout_init_set_15.txt};
    \addplot [only marks, color=NavyBlue, mark=*] table [y=Yredal15,x=y]{data/s3dis/s3dis_ReDAL_init_set_15.txt};
    \addplot [only marks, color=NavyBlue, mark=*] table [y=Yrand15,x=y]{data/s3dis/s3dis_random_init_set_15.txt};
    \addplot [only marks, color=NavyBlue, mark=*] table [y=Ysege15,x=y]{data/s3dis/s3dis_segment_entropy_init_set_15.txt};
    \addplot [only marks, color=NavyBlue, mark=*] table [y=Ysofc15,x=y]{data/s3dis/s3dis_softmax_confidence_init_set_15.txt};

    \end{axis}
\end{tikzpicture}
    \vspace{-35pt}
    \caption{\emph{Correlation between results of AL methods when using a random set and the closest in the DINO feature space~\cite{caron2021dino}.} Each point corresponds to a cycle of an AL method. Pearson correlation coefficient is 0.78. Results are produced on the S3DIS dataset~\cite{s3dis}.}
    \label{fig:dino-v2}
\end{figure}
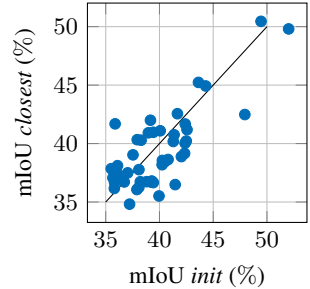

\begin{figure*}[ht!]
\centering
\includegraphics[width=0.95\linewidth]{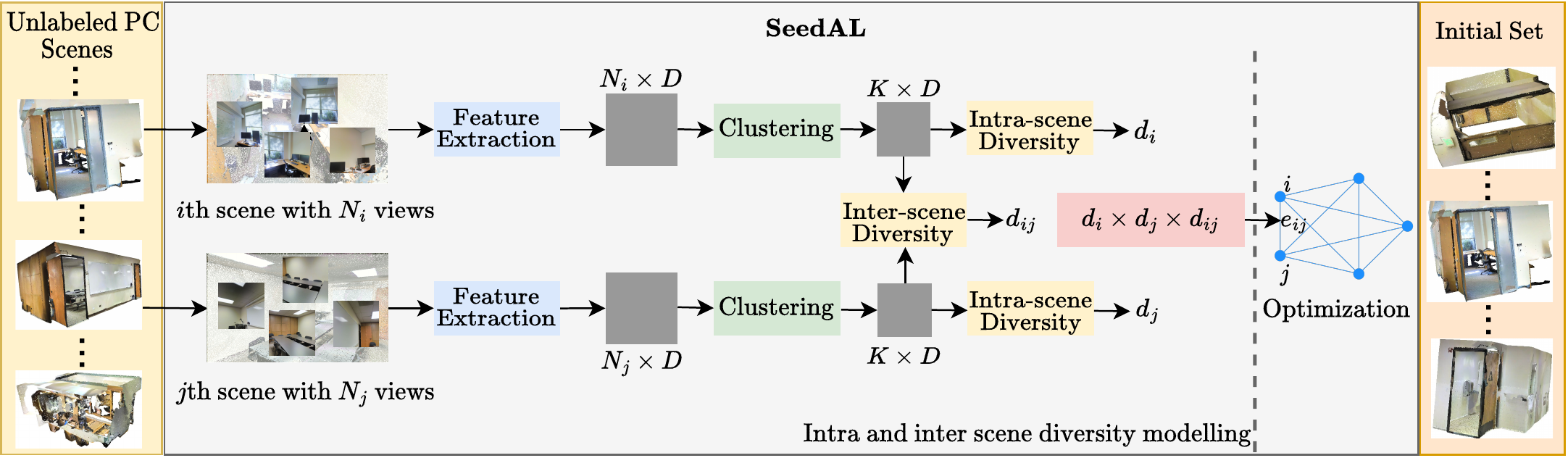}
\vspace{-5pt}
\caption{Overall processing pipeline of SeedAL.} 
\label{fig:model}
\vspace{-10pt}
\end{figure*}   

\subsection{Image features to characterize 3D scenes}
\label{sec:2Dfeatratherthan3Dfeat}

At first glance, it seems more natural and appropriate to directly use 3D features, as opposed to rely on extra 2D images. However, 3D backbones are not as versatile as 2D ones because of the huge domain gaps in 3D data, compared to picture data. 3D data require specific architectures and tuning due to the extreme variations of 3D point distribution, depending on the type of scenes (indoors vs outdoors, static vs dynamic) and on the kind of sensor used (photogrammetry, variants of depth cameras, lidars with widely differing acquisition patterns and resolutions).
Besides, current self-supervised 3D features have a lower quality than their 2D counterpart in the sense that self-supervised 3D features can be improved by distilling, in 3D, high-quality 2D image features of views of the same scenes~\cite{slidr}. In fact, the quality of self-trained 2D image features \cite{grill2020byol,gidaris2021obow,caron2021dino,bardes2022vicreg}, is now such that, even if self-trained on different datasets, it can be used for object discovery in new data \cite{lost, wang2022tokencut, ponimatlin2023uvos}. The best results have actually been achieved with DINO~\cite{caron2021dino}\OS{; \cite{yehuda2022coveringlens} also exploit DINO in their AL method}. 
There is no such result with pretrained 3D features yet. Last, 2D backbones are pretrained efficiently and on very large datasets, which boosts quality and favors generalizability, compared to the computation burden and practical size of pretraining data in 3D.

Here, rather than relying on a distillation from 2D to 3D as in \cite{slidr}, it is simpler and more efficient to directly use 2D image features to select corresponding 3D fractions of the dataset. 
Moreover, it makes our approach agnostic to the 3D backbone, which avoids having to handle the domain gaps of 3D data. Last, we can leverage very large image datasets, such as ImageNet, as it has been shown to transfer well to tasks on other image datasets~\cite{caron2021dino}; pretraining specifically on the images of the target dataset is not needed, and could even be detrimental if the dataset is not large enough.

\subsection{Relevance of DINO features}

To evaluate the effectiveness of DINO features, we first select a subset of the random seeds generated previously. \nermin{Next, we create the most similar seed \renaud{(in terms of DINO features) to} 
each selected random seed, and run active learning methods with all of the seeds. Finally, we calculate the correlation between \renaud{the} random seeds and their newly generated most similar correspondences. More specifically,} for each selected seed, denoted as $S$, we replace all scenes in each seed by identifying alternative scenes with the minimum distance in DINO space, and create a new seed denoted as $\hat{S}$.  
Formally, for each scene $i$, we compute the average feature $\Phi^i = \sum_{n=1}^{N_i} \phi(V^i_{N_i}) / N_i$ of all views. Then, we create $\hat{S}$ by selecting, for each scene $i \in S$, the scene $\argmin_{j \notin S} \| \Phi^j - \Phi^i \|_2$. 
Then,  we apply several active learning~(AL) methods to the newly created initial sets, and compare the results obtained from $S$ and $\hat{S}$.
Our analysis reveals a strong positive correlation between the two sets of results with a Pearson correlation coefficient of 0.78. The results indicate the applicability of leveraging DINO features for selecting the initial seed.

\section{SeedAL: Method}
\label{sec:method}

Our approach to seed selection posits that a good set should have two key characteristics: Firstly, each scene within the set should exhibit a sufficient level of internal diversity. Secondly, there should be a notable degree of diversity between the scenes themselves.
To identify the scenes that should be in the seed, we represent the unlabeled dataset as a fully connected graph $G$, in which each node represents a scene. The edge weights, denoted by $e_{ij}$, between two scenes $i$ and $j$, encapsulate both intra and inter-scene diversity measures. These measures are formally introduced below. 
By looking for a sub-graph that optimizes the sum of edge values while satisfying a budget constraint, we can find a seed with good diversity. We propose a linear optimization framework for the selection of this sub-graph and explain its details in the following \gilles{paragraphs}.
We illustrate our method in \autoref{fig:model}.
 
\subsection{Diversity measures}
\paragraph{Intra-scene diversity measure.} Given that a scene may contain near-duplicate views, e.g.,
captured from very similar viewpoints, it is important to eliminate redundancy to obtain representative features that accurately capture the characteristics of the scene. 
To this end, for each scene $i$, we cluster the set of view-features $\phi(V^i_1), \ldots, \phi(V^i_{N_i})$ in $K$ clusters using k-means. 
The resulting set of representative features, denoted by 
\gilles{$\Phi^i_1, \ldots, \Phi^i_K$}, contains $K$ distinctive features of the scene. 
To calculate the diversity within a scene $i$, we calculate the pairwise dissimilarities between the cluster centers \gilles{$\Phi^i_1, \ldots, \Phi^i_K$}: 
\gilles{$(1 - \Phi^i_k \cdot \Phi_{k'}^i)$ for all $k > k' \in \{1, \ldots, K\}$, where ``$\cdot$'' denotes the scalar product in $\mathbb{R}^D$.}
\gilles{Finally, we average all these pairwise dissimilarities to obtain the internal diversity $d_i$ of scene $i$:}
\begin{equation}
\label{eq:diversity}
   \gilles{d_i = \frac{2}{K(K-1)} \; \sum_{k=1}^{K} \sum_{k'=k+1}^{K} (1 - \Phi_{k}^i \cdot \Phi_{k'}^i) \>}.
\end{equation}

\paragraph{Inter-scene diversity measure.}
\gilles{Given two different scenes $i$ and $j$, we compute their inter-scene diversity $d_{ij}$ by relying on their respective cluster centers: $\Phi^i_1, \ldots, \Phi^i_K$ and $\Phi^j_1, \ldots, \Phi^j_K$. Specifically, we average the pairwise dissimilarities between all pairs of cluster centers:
\begin{equation}
\label{eq:interdiversity}
d_{ij} = \frac{1}{K^2} \sum_{k=1}^{K} \sum_{k'=1}^{K} (1 - {\Phi}_{k}^{i} \cdot {\Phi}_{k'}^{j}) \> .
\end{equation}
}

\paragraph{Combined intra and inter-scene diversity measure.}
\gilles{To combine the intra and inter-scene diversities, we simply compute $e_{ij} = d_{ij} \, d_{i} \, d_{j}$. Hence, scenes $i$ and $j$ in the graph $G$ are highly connected if they are both highly diverse internally \emph{and} mutually diverse.}

\subsection{Linear optimization for seed selection}

We cast our seed selection as the identification of a fully connected subgraph  of $G$ such that the sum of edge weights is maximized under the constraint of a budget $b$ for the initial annotations. Here, we define this budget in terms of the number of points to be annotated. The connected subgraph of $G$ could be identified by solving the following problem:
\begin{align}
\label{eq:quadoptim}
\underset{\{x_i\}_i}{\argmax} \quad 
\sum_{i=1}^M\sum_{j=i+1}^{M} e_{ij}x_ix_j \quad 
\text{s. t.} \quad \sum_{i=1}^{M} c_i x_i \le b \> ,
\end{align}
where the $x_i$ are boolean variables, indicating whether scene $i$ is selected or not.
$c_i$ the annotation cost for scene $i$, which in practice we take as the number of points in scene $i$.
The objective term $\sum_{i=1}^M\sum_{j=i+1}^M e_{ij}x_ix_j$, where $M$ is the number of scenes, permits us to select scenes with the highest diversity possible while the constraint $\sum_{i=1}^{M} c_ix_{i} \le b$ ensures that we \nermin{do not} exceed our annotation budget $b$. 

However, this is a quadratic problem and $M$ is large in practice. We can turn this problem into the linear problem
\vspace{-5mm}
\begin{align}
\label{eq:linearoptim}
\underset{\substack{\{y_{ij}\}_{i,j>i} \\ \{x_i\}_i}}{\argmax} \; \sum_{i=1}^{M}\sum_{j=i+1}^{M} e_{ij}y_{ij}  
 \text{ s. t. } 
\begin{cases}
\sum_{i=1}^{M} c_ix_{i} \le b \\ 
\quad y_{ij} \le x_{i} \\ 
\quad y_{ij} \le x_{j} \\ 
\end{cases}
\end{align}
by introducing the boolean variables $y_{ij}$. Because it is an optimization problem and thanks to the additional constraints, $y_{ij} = 1 \Leftrightarrow x_i = x_j = 1$. 
The advantage of this formulation is that it is much more efficient to solve when the number of scenes $M$ is large.

In practice, we observed that it can still be computationally challenging for very large $M$.  To address this issue, we first extract the top $L$ edges in $G$ with the highest weights $e_{ij}$ and collect all scenes on both ends on these edges. Then, we rebuild a smaller graph using only these scenes and apply the optimization to this reduced graph.
We use the solver from SCIP~\cite{BestuzhevaEtal2021ZR} to find the optimal solution.

\section{Experiments}
\label{sec:experiments}

\begin{figure*}[ht!]
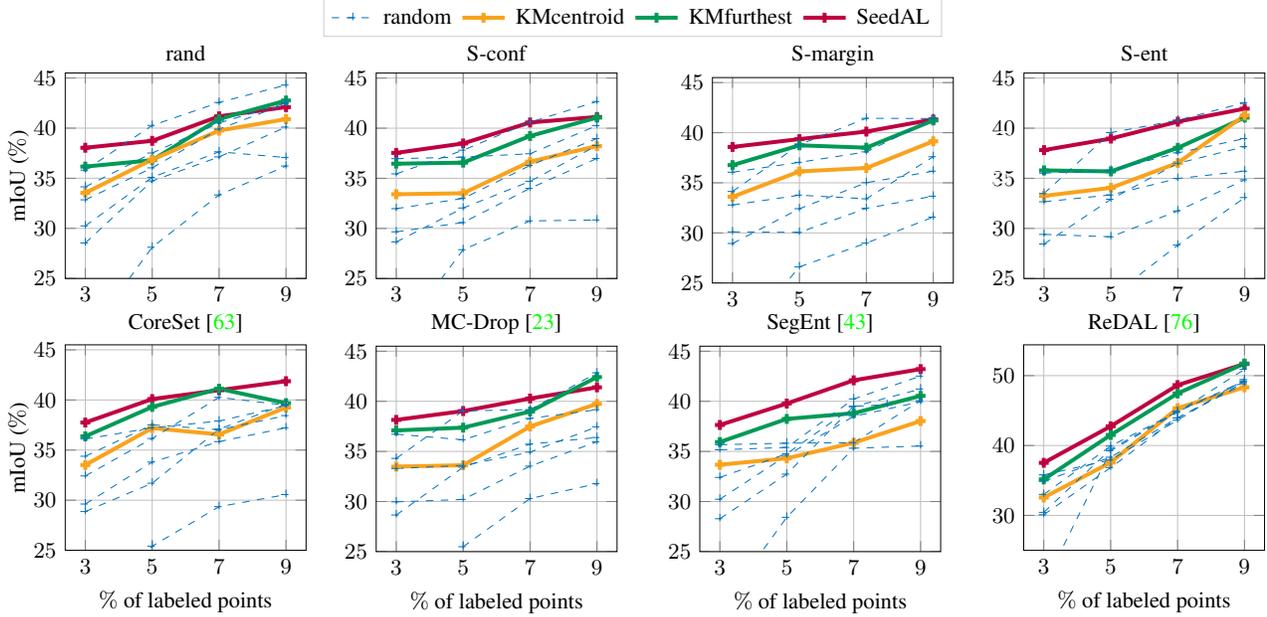

\centering
    \include{figs/tikz/all_methods_s3dis}
    \caption{\textbf{AL results on S3DIS.} Comparison of active learning results when using \ours vs several baselines.  }
    \label{fig:sota-s3dis}
    \vspace{-5pt}
\end{figure*}

\OS{We present in this section the experimental protocol that we use to evaluate our method \ours and discuss results. We also provide an ablation study.}

\subsection{Experimental setup}
\label{sec:setup}

\paragraph{Active learning strategies.}
We test our method with diverse active learning strategies that exploit either the model uncertainty or enforce diversity, 
that are either scene-based or region-based. In particular, we compare to a random selection of scenes~(\random), the diversity-based method CoreSet~\cite{coreset}~(\coreset) and the uncertainty-based methods Softmax Confidence~\cite{softmax}~(\softconf), Softmax Margin~\cite{softmax}~(\softmargin), Softmax Entropy~\cite{softmax}~(\softentropy), MC-Dropout~\cite{mcdropout}~(\mcdropout) and Segment Entropy~\cite{seg_ent}~(\segent). We also consider the region-based method 
ReDAL~\cite{wu2021redal}~(\redal).

\paragraph{Seeding baselines.} We compare \ours to two clustering-based baselines inspired from~\cite{yuan2020cold}. For this purpose, for each scene $i$, we first compute the average feature $\Phi^i = \sum_{n=1}^{N_i} \phi(V^i_{N_i}) / N_i$ of all views. 
The first simple baseline, \KMfirst,
starts with a clustering of the features $\Phi^1, \ldots, \Phi^M$ into $K$ clusters using k-means. Then, for each cluster, we search the scene $i$ whose feature $\Phi^i$ is the closest to the cluster centroid, we select these $K$ scenes, and repeat this search process until we reach the annotation budget. The most costly samples are then removed iteratively until the annotation budget is satisfied.
We also consider a variant of \KMfirst, called \KMsecond, inspired by the core-set \cite{coreset} selection. It starts as in \KMfirst with a selection of the $K$ scenes closest to each cluster centroid. Then, 
we continue spending the annotation budget by selecting, iteratively, the scene $j$ whose feature $\Phi^j$ is the farthest away from all cluster centroids, to favor a selection of diverse scenes. 
We also compare to the natural 
baseline \emph{random}.

\paragraph{Datasets.} 
Following \cite{wu2021redal}, we focus on the semantic segmentation task and evaluate \ours on two datasets representative of indoor and outdoor scenes, namely S3DIS~\cite{s3dis} and SemanticKITTI~\cite{semantickitti}. 
S3DIS is composed of 271 scenes extracted from 6 major indoor areas; for each scene is provided a dense point cloud with  
color.  
As common practice, we provide AL results 
on the Area\,5 validation set, and use the rest for training. 
We also evaluate on SemanticKITTI, a large-scale autonomous driving dataset composed of 22 driving sequences, totaling 43,552 point clouds, each accompanied with images. 
Following the official protocol, we evaluate on the validation split (seq 08) and train the models on the entire official training split (seq 00-07 and 09-10).

\paragraph{Network architecture and evaluation metric.} Following previous works~\cite{wu2021redal}, we consider SPVCNN~\cite{spvcnn}, that is based on point-voxel CNN and achieves good results both on indoor and outdoor scenes. We evaluate the semantic segmentation task in all experiments with the mIoU metric.

\paragraph{Technical details.}
In all our experiments, we set $K$ as the number of the classes for each dataset: 13 for S3DIS, 19 for  
SemanticKITTI., 
$L$ is set as 100 for S3DIS and 1000 for SemanticKITTI. 
On S3DIS, the RGB images depicting each scene constitute our views. We use the class token extracted at the last layer of the DINO-pretrained ViT-B/8 as our view features. On SemanticKITTI, each scene is depicted by a single image frame. To account for multi-object images in outdoor scenes, we consider that each $8 \ssp\times 8$ patches of a frame constitute one view and we take the corresponding patch feature at the last layer of the ViT-B/8 as the view feature.
Also, as there is a significant number of redundant and highly similar scenes in SemanticKITTI, we use a greedy algorithm to sparsify the sequences. The details of this sparsification are in the supplementary material.

\begin{figure*}[ht!]
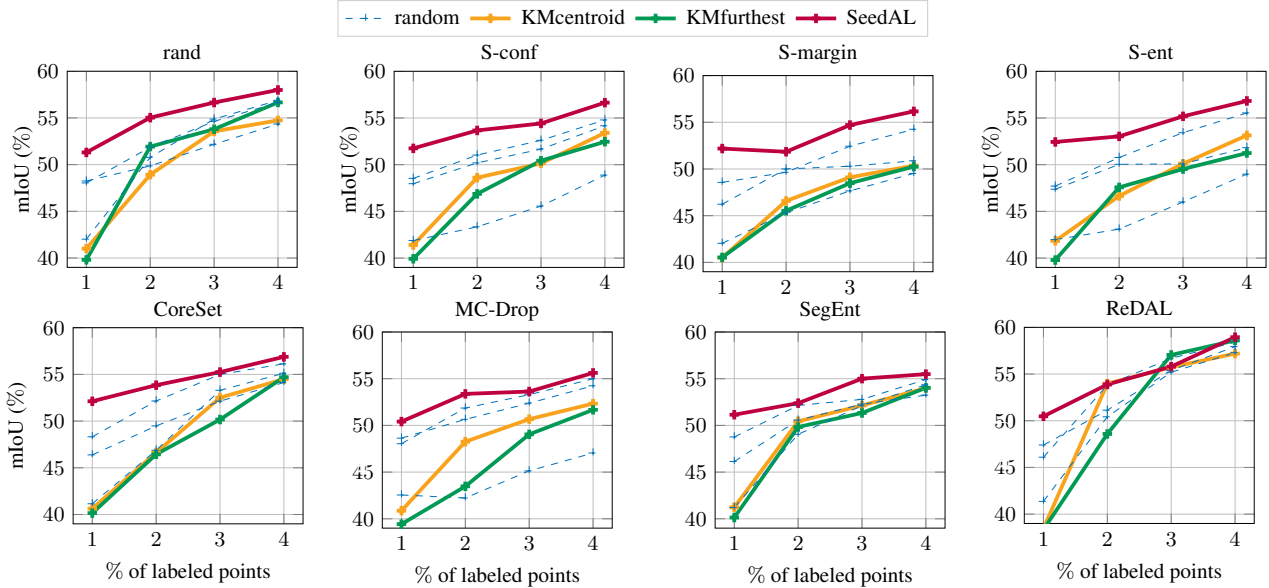

    \include{figs/tikz/all_methods_kitti}
        \vspace{-5pt}
    \caption{\textbf{AL results on SemanticKiTTI.} Comparison of active learning results when using \ours vs several baselines. }
    \label{fig:sota-kitti}
    \vspace{-10pt}
\end{figure*}

\subsection{Indoor evaluation}
We evaluate here our method on S3DIS.
We present in \autoref{fig:sota-s3dis} the results obtained with our AL seeding strategy, \ours, and  with the considered baselines. 
First, we observe that using \ours boosts significantly all 8 AL methods evaluated over randomly selected seeds (dashed blue curves). 
\gilles{\ours is consistently at the top of the graphs while random seeds yields, overall, unstable results.}
Regarding our baselines, \KMsecond achieves consistently better performances than \KMfirst over all AL strategies. \gilles{We explain this result by the fact that \KMsecond contains more diverse scenes than \KMfirst.} We observe that for certain methods, \KMsecond achieves similar levels of performance as SeedAL. However, it is important to note that SeedAL is more robust and it consistently obtains strong performance in both early and late active learning iterations.

\paragraph{\OS{Making a good first impression.}}
We highlight here the boost obtained with \ours at the \emph{very first cycle} (when no AL has been applied yet). We observe that with \ours, results are \nermin{+1.9 pt} over the best random seed, \nermin{+7.8 pt} v.s.\ the average of random seeds and \nermin{+18.8 pt} when comparing to the most unlucky seed. We surpass also our two baselines \KMfirst and \KMsecond by \nermin{4.9 pt} and \nermin{2.4 pt}, respectively. 
With only $3\%$ of data well-selected \gilles{thanks to \ours}, we obtain better results than with 4 out of 5 random seeds with $5\%$, or even $7\%$ (when employing \softconf, \softmargin, \softentropy, \coreset and \mcdropout), of the data. These results prove that our seeding selection is a much better strategy than betting on the quality of the random seed.

\nermin{We also compare our method with SSDR-AL~\cite{ssdr}. Differently than other AL methods, SSDR-AL's initialization picks superpoints to label, \renaud{rather than full scenes}. As our current formulation selects scenes rather than regions, it cannot directly be applied \renaud{in this case}. \renaud{The only way we can compare quantitatively} 
with~\cite{ssdr} is on S3DIS regarding the \renaud{proportion of points} 
to label \renaud{in order to reach the same level as} 
90\% of full supervision: we boost ReDAL to outperform SSDR-AL, only requiring 9\% labeled points instead of 11.7\%, as shown in \autoref{tab:ssdr}.
 
\begin{center}
\vspace*{-1.0mm}\def\tabcolsep{0.8mm}
\resizebox{0.47\textwidth}{!}{
\begin{tabular}{ l|@{~~}cccc} 
 \hline
 Method & SSDR & ReDAL & ReDAL & ReDAL \\ 
 (init.) & (rand) & (rand) & (ours-DINO) & (ours-MoCo)\\ 
 \hline
\% labeled pts ($\downarrow$) & 11.7 \% & 13\% & {11\%} & \bf 9\%\\
 \hline
\end{tabular}
}
\vspace*{-2mm}
\captionof{table}{Comparing SSDR's initialization with SeedAL in terms of \% of pts to label to get to 90\% of full supervision. }
\label{tab:ssdr}
\end{center} 
}

\begin{figure*}[ht!]
    \centering
    \resizebox{0.95\textwidth}{!}{
\begin{tabular}{cccc}
    \multicolumn{4}{c}{
    \centering
    \begin{tikzpicture}
    \begin{customlegend}[legend columns=2,legend style={align=center, column sep=.4ex, nodes={scale=0.9, transform shape}, draw=white!80!black},
        legend entries={
                        \coreset,
                        \redal,
                        }]
        \addlegendimage{color=black, mark=+}
        \addlegendimage{color=black, dashed, mark=+} 
        \end{customlegend}
    \end{tikzpicture}
    }
    \vspace{-5pt}
    \\
\centering
\centering
\begin{tikzpicture}
    \tikzstyle{every node}=[font=\small]
    \begin{axis}[
        ymax=60,
        width=4.9cm,
        height=4.5cm,
        font=\small,
        xtick={3,5,7,9},
        ylabel=mIoU ($\%$),
        xlabel=$\%$ of labeled points,
        ylabel shift=-5 pt,
        xlabel shift=-5 pt,
        label style={font=\small},
        tick label style={font=\small},
        title style={yshift=-27.ex,},
        title=(a),
        legend pos=north west,
        grid=major,
        legend cell align={left},
        legend style={fill opacity=0.8, draw opacity=1, text opacity=1, draw=white!80!black, nodes={scale=0.8, transform shape}},
    ]

    \addplot [color=Orange, thick, mark=+] table [y=y,x=x]{data/s3dis/s3dis_coreset_init_set_ablation_internal_dsim.txt};
    \addlegendentry{intra-div.};
    \addplot [color=Orchid, thick, mark=+] table [y=y,x=x]{data/s3dis/s3dis_coreset_init_set_ablation_internal_sim.txt};    
    \addlegendentry{intra-sim.};

    \addplot [color=Orange, thick, dashed, mark=+] table [y=y,x=x]{data/s3dis/s3dis_redal_init_set_ablation_internal_dsim.txt};
    \addplot [color=Orchid, thick, dashed, mark=+] table [y=y,x=x]{data/s3dis/s3dis_redal_init_set_ablation_internal_sim.txt};    
    
    \end{axis}
\end{tikzpicture}

    & 
 \centering
\begin{tikzpicture}
    \tikzstyle{every node}=[font=\small]
    \begin{axis}[
        ymax=60,
        width=4.9cm,
        height=4.5cm,
        font=\small,
        xtick={3,5,7,9},
        xlabel=$\%$ of labeled points,
        label style={font=\small},
        tick label style={font=\small},
        xlabel shift=-5 pt,
        title=(b),
        title style={yshift=-27.ex,},
        legend pos=north west,
        grid=major,
        legend cell align={left},
        legend style={fill opacity=0.8, draw opacity=1, text opacity=1, draw=white!80!black, nodes={scale=0.8, transform shape}},
    ]

    \addplot [color=Orange, thick, mark=+] table [y=y,x=x]{data/s3dis/s3dis_coreset_init_set_ablation_internal_dsim_w_clustering.txt};
    \addlegendentry{cls. feat.};
    \addplot [color=Orchid, thick, mark=+] table [y=y,x=x]{data/s3dis/s3dis_coreset_init_set_ablation_internal_dsim.txt};    
    \addlegendentry{feats.};

    \addplot [color=Orange, dashed, thick, mark=+] table [y=y,x=x]{data/s3dis/s3dis_redal_init_set_ablation_internal_dsim_w_clustering.txt};
    \addplot [color=Orchid, dashed, thick, mark=+] table [y=y,x=x]{data/s3dis/s3dis_redal_init_set_ablation_internal_dsim.txt};    
    
    \end{axis}
    \end{tikzpicture}

    & 
\centering
\begin{tikzpicture}
    \tikzstyle{every node}=[font=\small]
    \begin{axis}[
        ymax=60,
        width=4.9cm,
        height=4.5cm,
        font=\small,
        xtick={3,5,7,9},
        xlabel=$\%$ of labeled points,
        label style={font=\small},
        tick label style={font=\small},
        xlabel shift=-5 pt,
        title=(c),
        title style={yshift=-27.ex,},
        legend pos=north west,
        grid=major,
        legend cell align={left},
        legend style={fill opacity=0.8, draw opacity=1, text opacity=1, draw=white!80!black, nodes={scale=0.8, transform shape}},
    ]
    \addplot [color=Orange, mark=+] table [y=y,x=x]{data/s3dis/s3dis_coreset_init_set_ablation_external_dsim_w_clustering.txt};
    \addlegendentry{inter-div.};
    \addplot [color=Orchid, mark=+] table [y=y,x=x]{data/s3dis/s3dis_coreset_init_set_ablation_external_sim_w_clustering.txt};
    \addlegendentry{inter-sim.};
    
    \addplot [color=Orange, thick, dashed, mark=+,forget plot] table [y=y,x=x]{data/s3dis/s3dis_ReDal_init_set_ablation_external_dsim_w_clustering.txt};
    \addplot [color=Orchid,  thick, dashed, mark=+, forget plot] table [y=y,x=x]{data/s3dis/s3dis_ReDal_init_set_ablation_external_sim_w_clustering.txt};
    
    \end{axis}
\end{tikzpicture}

    &
  \centering
\begin{tikzpicture}
    \tikzstyle{every node}=[font=\small]
    \begin{axis}[
        ymax=60,
        width=4.9cm,
        height=4.5cm,
        font=\small,
        xtick={3,5,7,9},
        xlabel=$\%$ of labeled points,
        label style={font=\small},
        tick label style={font=\small},
        xlabel shift=-5 pt,
        title=(d),
        title style={yshift=-27.ex,},
        legend pos=north west,
        grid=major,
        legend cell align={left},
        legend style={fill opacity=0.8, draw opacity=1, text opacity=1, draw=white!80!black, nodes={scale=0.8, transform shape}},
    ]

    \addplot [color=Orange,  thick, mark=+] table [y=y,x=x]{data/s3dis/s3dis_coreset_our_full_method.txt};
    \addlegendentry{SeedAL};
    \addplot [color=Orchid,  thick, mark=+] table [y=y,x=x]{data/s3dis/s3dis_coreset_init_set_ablation_internal_dsim_w_clustering.txt};
    \addlegendentry{intra-div.};
    \addplot [color=ProcessBlue,  thick, mark=+] table [y=y,x=x]{data/s3dis/s3dis_coreset_init_set_ablation_external_dsim_w_clustering.txt};
    \addlegendentry{inter-div.};

    \addplot [color=Orange, mark=+, dashed] table [y=y,x=x]{data/s3dis/s3dis_redal_our_full_method.txt};
    \addplot [color=Orchid, mark=+, dashed] table [y=y,x=x]{data/s3dis/s3dis_redal_init_set_ablation_internal_dsim_w_clustering.txt};
    \addplot [color=ProcessBlue, mark=+, dashed] table [y=y,x=x]{data/s3dis/s3dis_redal_init_set_ablation_external_dsim_w_clustering.txt};
    
    \end{axis}
\end{tikzpicture}

\end{tabular}
}
    \vspace{-21pt}
    \caption{%
    \emph{Ablation study.} We evaluate here results obtained with different seeding strategies. (a) Seeds made of scenes with high intra-diversity (intra-div.) or high intra-similarity (intra-sim.). (b) Seeds selected with two different intra-diversity metrics:
    view features (feats.)
    or computed after clustering the view features (cls. feat.). (c) Seeds made of scenes with high inter-diversity (inter-div.) or high inter-similarity (inter-sim.). (d) Seeds selected with \ours, considering only inter-diversity (inter-div.) or intra-diversity (intra-div.).
    \vspace{-10pt}
    }
    \label{fig:ablation}
\end{figure*}
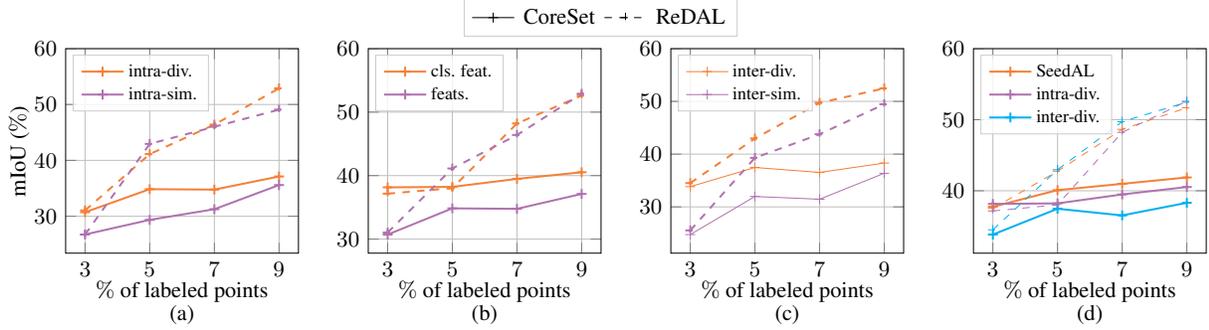

\subsection{Outdoor evaluation}

We now focus on the challenging SemanticKitti dataset, which depicts highly redundant urban images. We present results 
in  \autoref{fig:sota-kitti}.
We also show results with the baseline \random, which is known to surpass all AL methods except the region-based \redal. 

\paragraph{Boosting active learning.}
First, we observe that \ours significantly surpasses all seeding baselines, either random or diversity-based. Second, it
achieves better results compared to any randomly sampled seed with all AL methods and at all cycles but one (with \redal). 
\gilles{\ours yields an improvements of \nermin{3.3 pt} and \nermin{9.7 pt} in the first cycle over the best and worst random seed, respectively.}
We also observe that the results obtained with \KMfirst and \KMsecond are rather disappointing: the AL results are falling below the worst random seed at the first and second cycles. 
Surprisingly, ReDAL reaches better results with \KMsecond when using $3\%$ of the dataset. 
However, our seeding strategy appears to be \emph{agnostic} to the AL method and boost all of them consistently. Overall, the best results are again obtained with our seeding strategy and \redal.

SeedAL is especially effective for large-scale datasets, enabling scene-based AL methods to achieve almost 90\% of the fully supervised performance using just 4\% labeled points. This is comparable to the region-based method ReDAL which achieves 93\% of the fully supervised performance. SeedAL bridges the gap between complex and costly methods and simpler AL methods on the large-scale SemanticKITTI dataset.

\paragraph{Improving random selection.} \random is known to be a hard-to-beat baseline on this dataset \cite{wu2021redal}.
We observe that \gilles{thanks to \ours, } 
the performance of \textit{\random} goes 
\gilles{as high as 58.0\%, v.s.\ 58.9\% for the best considered method (\redal).} 
The baseline \random is not itself an active learning strategy --- given that scenes are randomly sampled ---but these results show that a good seeding strategy impacts greatly all results, even those of the random baseline. 

\nermin{We also compare our method with LiDAL~\cite{hu2022lidal} on SemanticKITTI. LiDAL is an AL method specifically developed for 3D LiDAR semantic segmentation. As it is shown in \autoref{tab:lidal}, LiDAL benefits more from SeedAL than random selection.
It gets 90\% of full supervision's mIoU (58.0\%) with 2.1\% labels, vs 2.6\% with a random seed. 
 
\begin{center}
\vspace*{-1.0mm}\def\tabcolsep{1.5mm}
 \resizebox{0.9\columnwidth}{!}{
\begin{tabular}{ l|ccccc } 
 \hline
 Seeding method & Init (1\%) & 2\%  &  3\% & 4\% & 5\%\\ 
 \hline
 \random~\cite{hu2022lidal} & 48.8 & 57.1 & 58.7 & 59.3 & 59.5 \\  
 SeedAL (ours) & \textbf{52.6} & \textbf{57.8} & \textbf{59.3} & \textbf{60.3} & \textbf{60.6} \\
 \hline
\end{tabular}
}
\vspace*{-2mm}
\captionof{table}{LiDAL~\cite{hu2022lidal} with random \renaud{or SeedAL's seeds}.}
\label{tab:lidal}
\end{center}
 }

\subsection{Ablation experiments}

We study the relevance of the different components used to construct \ours. We conduct experiments on S3DIS with all the active learning methods described in Sec.~\ref{sec:setup}. Due to space constraint, we report CoreSet and ReDAL results and include the others in the supplementary material.

\paragraph{Intra-scene diversity.}
We highlight the relevance of selecting scenes with high intra-scene diversity instead of high intra-scene similarity. We compute a diversity and a similarity score per scene. Then we select the top scenes according to each metric until exhausting our annotation budget. The diversity and similarity scores are respectively computed by averaging ($1 - \phi(V^i_{N_i}) \cdot \phi(V^i_{N_j})$) and ($\phi(V^i_{N_i}) \cdot \phi(V^i_{N_j})$) between all views of a scene. Results in \autoref{fig:ablation}~(a) show that 
using the intra-scene diversity yields the best results.

\paragraph{Clustering views.}
We show that computing the dissimilarity score from the cluster centroids as in \eqref{eq:diversity} is a better strategy than computing it as in the previous paragraph.
Performances obtained with both strategies are presented in \autoref{fig:ablation}~(b). Clustering removes redundant features and yields a better estimate of the scene diversity, hence selection of a significantly better seed as demonstrated by the results for 3\% of labeled points.

\paragraph{Inter-scene diversity.}
We justify why we select scenes with large inter-scene diversity rather than scenes with large inter-scene similarity. 
In a first experiment, we replace the edge weight $e_{ij}$ in problem \eqref{eq:linearoptim} by $d_{ij}$, the inter-scene dissimilarity of \eqref{eq:interdiversity}, and, in a second experiment, by the inter-scene similarity, obtained by averaging (${\Phi}_{k}^{i} \cdot {\Phi}_{k'}^{j}$) over all pairs $(k, k')$.  \autoref{fig:ablation}~(c) shows that selecting scenes with high inter-scene dissimilarity yields the best seeds.

\paragraph{Combined intra and inter-scene diversity.}
Finally, we show the interest of combining the intra and inter-scene diversity, i.e., using \ours, in \autoref{fig:ablation}~(d). In this figure, we compare the performance of seeds selected by solving equation~\eqref{eq:linearoptim} with $e_{ij} = d_id_jd_{ij}$ (\ours), or $e_{ij} = d_{ij}$ (inter-div). We also show the performance of seeds made of scenes with top $d_i$ (intra-div) intra-diversity.

\begin{figure}[t!]
\centering
    \resizebox{\columnwidth}{!}{
\begin{tabular}{cc}
    \multicolumn{2}{c}{
    \centering
    \begin{tikzpicture}
    \begin{customlegend}[legend columns=3,legend style={align=center, column sep=.4ex, nodes={scale=0.8, transform shape}, draw=white!80!black},
        legend entries={ 
                        \depthcontrast,
                        \also,
                        \segcontrast,
                        \moco,
                        \dino,
                        random,
                        }]
        \addlegendimage{ultra thick, color=YellowOrange, mark=+}
        \addlegendimage{ultra thick, color=Orchid, mark=+}
        \addlegendimage{ultra thick, color=Cyan, mark=+}
        \addlegendimage{ultra thick, color=ForestGreen, mark=+} 
        \addlegendimage{ultra thick, color=purple,  mark=+}
        \addlegendimage{color=RoyalBlue, dashed, mark=+} 
        \end{customlegend}
    \end{tikzpicture}
    }
    \vspace{-12pt}
    \\

\begin{tikzpicture}[baseline={([yshift=\baselineskip]current bounding box.north)}]
  \tikzstyle{every node}=[font=\small]
    \begin{axis}[
        ymin=30,
        ymax=45.5,
        width=4.6cm,
        height=4.2cm,
        font=\footnotesize,
        xtick={3,5,7,9},
        ylabel=mIoU ($\%$),
        label style={font=\small},
        tick label style={font=\small},
        title=\softmargin,
        title style={yshift=-1.ex,},
        legend pos=south east,
        grid=major,
        legend style={nodes={scale=0.8, transform shape}},
        legend cell align={left}
    ]
     
    \addplot[ultra thick, color=YellowOrange, mark=+]  table [y=y,x=x]{data/rebuttal/s3dis_depthcontrast_softmax_margin.txt};
     \addlegendentry{\depthcontrast};
     
    \addplot[ultra thick, color=ForestGreen, mark=+]  table [y=y,x=x]{data/rebuttal/s3dis_mocov3_softmax_margin.txt};
    \addlegendentry{\moco};

     \addplot[ultra thick, color=purple, mark=+]  table [y=y,x=x]{data/s3dis/s3dis_softmax_margin_our_full_method.txt};
     \addlegendentry{\dino}; 

    \addplot[color=RoyalBlue, dashed, mark=+]  table [y=y,x=x]{data/s3dis/s3dis_softmax_margin_random-mean.txt};
     \addlegendentry{random};
     
    \legend{};
    
    \end{axis}
\end{tikzpicture} 
&
\begin{tikzpicture}[baseline={([yshift=\baselineskip]current bounding box.north)}]
  \tikzstyle{every node}=[font=\small]
    \begin{axis}[
        ymin=30,
        ymax=62.5,
        width=4.6cm,
        height=4.2cm,
        font=\footnotesize,
        xtick={3,5,7,9},
        label style={font=\small},
        tick label style={font=\small},
        title=\redal,
        title style={yshift=-1.ex,},
        legend pos=south east,
        grid=major,
        legend style={nodes={scale=0.8, transform shape}},
        legend cell align={left}
    ]
     
     \addplot[ultra thick, color=YellowOrange, mark=+]  table [y=y,x=x]{data/rebuttal/s3dis_depthcontrast_ReDAL.txt};
     \addlegendentry{\depthcontrast};
     
     \addplot[ultra thick, color=ForestGreen, mark=+]  table [y=y,x=x]
     {data/rebuttal/s3dis_mocov3_ReDAL.txt};
     \addlegendentry{\moco};

     \addplot[ultra thick, color=purple, mark=+]  table [y=y,x=x]
     {data/s3dis/s3dis_ReDAL_our_full_method.txt};
     \addlegendentry{\dino};

    \addplot[color=RoyalBlue, dashed, mark=+]  table [y=y,x=x]{data/s3dis/s3dis_ReDAL_random-mean.txt};
    \addlegendentry{random};
     
    \legend{};
    \end{axis}
\end{tikzpicture} 
\\
\begin{tikzpicture}[baseline={([yshift=\baselineskip]current bounding box.north)}]
  \tikzstyle{every node}=[font=\small]
    \begin{axis}[
        ymin=39,
        ymax=60,
        width=4.6cm,
        height=4.2cm,
        font=\footnotesize,
        xtick={1,2,3,4},
        ylabel=mIoU ($\%$),
        xlabel=$\%$ of labeled points,
        label style={font=\small},
        tick label style={font=\small},
        title=\segent,
        title style={yshift=-1.ex,},
        legend pos=south east,
        grid=major,
        legend style={nodes={scale=0.8, transform shape}},
        legend cell align={left}
    ]
     
    \addplot[ultra thick, color=Orchid, mark=+]  table [y=y,x=x]{data/rebuttal/kitti_also_segment_entropy.txt};
     \addlegendentry{\also};
     
    \addplot[ultra thick, color=ForestGreen, mark=+]  table [y=y,x=x]{data/rebuttal/kitti_mocov3_segment_entropy.txt};
     \addlegendentry{\moco};

     \addplot[ultra thick, color=purple, mark=+]  table [y=y,x=x]{data/kitti/kitti_segment_entropy_our_full_method.txt};
     \addlegendentry{\dino};  

    \addplot[color=RoyalBlue, dashed, mark=+]  table [y=y,x=x]{data/kitti/kitti_segment_entropy_rand_mean.txt};
     \addlegendentry{random};

    \addplot[ultra thick, color=Cyan, mark=+]  table [y=y,x=x]{data/rebuttal/kitti_segcont_segment_entropy.txt};
    \addlegendentry{\segcontrast}; 
     
    \legend{};
    
    \end{axis}
\end{tikzpicture} 
&
  \begin{tikzpicture}[baseline={([yshift=\baselineskip]current bounding box.north)}]
    \tikzstyle{every node}=[font=\small]
    \begin{axis}[
        ymin=39,
        ymax=60,
        width=4.6cm,
        height=4.2cm,
        font=\footnotesize,
        xtick={1,2,3,4},
        xlabel=$\%$ of labeled points,
        ylabel shift=-5 pt,
        label style={font=\small},
        tick label style={font=\small},
        title=\coreset,
        title style={yshift=-1.ex,},
        legend pos=south east,
        grid=major,
        legend style={nodes={scale=0.8, transform shape}},
        legend cell align={left}
    ]

    \addplot[ultra thick, color=Orchid, mark=+]  table [y=y,x=x]{data/rebuttal/kitti_also_coreset.txt};
     \addlegendentry{\also};
     
    \addplot[ultra thick, color=ForestGreen, mark=+]  table [y=y,x=x]{data/rebuttal/kitti_mocov3_coreset.txt};
     \addlegendentry{\moco};

     \addplot[ultra thick, color=purple, mark=+]  table [y=y,x=x]{data/kitti/kitti_coreset_our_full_method.txt};
     \addlegendentry{\dino};

    \addplot[color=RoyalBlue, dashed, mark=+]  table [y=y,x=x]{data/kitti/kitti_coreset_rand_mean.txt};
     \addlegendentry{random};

    \addplot[ultra thick, color=Cyan, mark=+]  table [y=y,x=x]{data/rebuttal/kitti_segcont_coreset.txt};
    \addlegendentry{\segcontrast};
     
    \legend{};
\end{axis}
\end{tikzpicture} 

\\    
\end{tabular}

}
    \vspace{-15pt}
    \caption{\nermin{SeedAL results on S3DIS (first row) and SemanticKITTI (second row) using features from DepthContrast, SegContrast, ALSO, \moco, DINO. `random' is an average over the random seeds.} 
    \vspace{-6pt}
    }
    \label{fig:other-ss-feats-sh}
\end{figure}
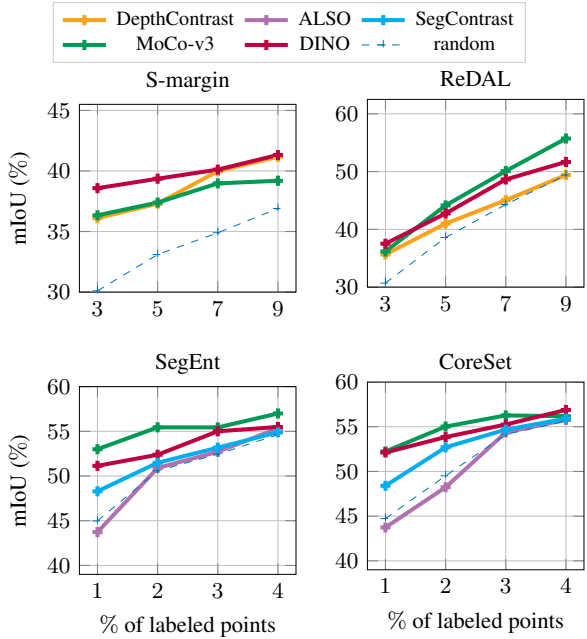

\nermin{Next we show the effect of different 2D and 3D self-supervised features to construct SeedAL. Due to space constraint, we report results from two different methods for each dataset and include the rest in supplementary material. 

\paragraph{\ours with different 2D 
features.}\!\! We experimented with self-supervised 2D features from \moco~\cite{chen2021mocov3} as an  alternative to DINO. The results are reported in \autoref{fig:other-ss-feats-sh}. On SemanticKITTI, SeedAL with \moco performs on par or slightly better than \renaud{SeedAL} with DINO, but is slightly inferior to  \renaud{SeedAL with} DINO on S3DIS, except for ReDAL. DINO and \moco are pretrained on ImageNet, which is very different from S3DIS and Semantic\-KITTI; it shows the foremost generalizability of these features.

\paragraph{\ours with pure 3D features.}\!\!  We \renaud{also experimented with self-supervised 3D features:} 
DepthContrast~\cite{depthcontrast} pretrained on ScanNet for indoor RGBD (no model is available for outdoors), \renaud{as well as} SegContrast~\cite{segcontrast} and ALSO~\cite{boulch2023also}
pretrained on Semantic\-KITTI.
To adapt \ours to work with 3D features, we simply replace the set of 2D image  features of a scene with 3D point features obtained from the point cloud of the same scene\renaud{, which we thus average to obtain whole-scene features}. 
Results are illustrated in \autoref{fig:other-ss-feats-sh}:
DepthContrast is only slightly below 2D features, SegContrast is not as good but still better than rand, while ALSO underperforms, likely because it is not contrastive. Note that there is no pretrained 3D network that works both for indoors and outdoors, contrary to 2D-based models.}

\section{Conclusion}
\label{sec:conclusion}
In this work, we have shown the influence of seeds on the performance of AL methods for point clouds, and proposed a method for efficiently selecting a seed that yields to good performance. Our approach works out of the box: it is agnostic to the AL method and to the 3D backbone used for the downstream task, and it does not require training on the dataset. We hope it will help scaling up point cloud semantic segmentation for practical applications.

\section*{Acknowledgments}
We thank Samet Hicsonmez for insightful discussions. This work was granted access to HPC resources
of IDRIS under
\renaud{GENCI allocations}  
2023-AD011013267R1,
2023-AD011013262R1 and 2023-AD011013413R1.  


\newpage

{\small
\bibliographystyle{ieee_fullname}
\bibliography{my_bibliography}
}

\newpage
\clearpage
\appendix

\section*{Supplementary material -- Overview}

\nermin{
This supplementary material is organized as follows.
\vspace{-3pt}
\begin{itemize}

\item  We explain our sparsification algorithm, which is  
used to eliminate redundant 
scenes in SemanticKITTI~\cite{semantickitti} (Section~\ref{sec:sparsification});

\item We provide further details regarding model training for comparing active learning (AL) methods  (Section~\ref{sec:implementation});

\item We present all the quantative results in a table obtained from all AL methods on S3DIS and SemanticKITTI (Section~\ref{sec:big_table});

\item Finally, we report the remaining results of our ablation experiments; (i) component analyses results obtained from  AL methods \random, \mcdropout, \softconf, \softmargin, \softentropy and \segent, (ii) results obtained from additional 2D (MoCo-v3) and 3D (DepthContrast, SegContrast, ALSO) features for all AL methods (Section~\ref{sec:ablation_exps}); 

\end{itemize}
}

\vspace{-8pt}
\section{Sparsification of SemanticKITTI}
\label{sec:sparsification}

SemanticKITTI consists of sequences of frames sampled at 10\,Hz. Consequently, there is a high similarity between successive frames, which are thus somehow redundant.
To address this issue and improve scalability, we use a greedy algorithm to sparsify the SemanticKITTI dataset. 

For each sequence, we begin with the first frame, use it as reference, and calculate its similarity with  subsequent frames. 
We then eliminate any subsequent frame whose similarity with the first frame is above a threshold. 
The first subsequent frame falling below the threshold is then itself used as a new reference, and the process continues for all frames in the sequence. With this simple sparsification, we increase the scalability of the dataset and reduce computational requirements for downstream processing. 

The similarities are computed again using global DINO features for each frame. 
We set a threshold of 0.75 on the cosine similarity. Our algorithm reduces the size of SemanticKITTI by 95\%.

\vspace{-5pt}
\section{Implementation and experiment details}
\label{sec:implementation}

Our AL seeding method (\ours) is implemented using PyTorch. 
We run S3DIS~\cite{s3dis} experiments on a single V100 GPU with a batch size of 4. We perform the training of segmentation networks for 
CoreSet, S-conf, S-margin, S-ent, MC-drop and SegEnt on SemanticKITTI using 2 A100 GPUs with a batch size of 32. The training of networks when using \redal, a region-based method, is about 5~times longer than when using scene-based methods because more point clouds need to be processed.
\vspace{-0.5em}

\nermin{\paragraph{Running time includes:} using a pretrained model to create the features (90\,ms/image on a V100 GPU), clustering and sorting candidates (negligible time), and extracting the best ones within the budget by linear optimization ($<$\,1\,min for S3DIS, $<$\,5\,min for SemanticKITTI).}

\vspace{-6pt}
\section{Quantitative Results}
\label{sec:big_table}

To make it easier to compare performance quantitatively across papers, we 
report in \autoref{tab:quantitative} the detailed quantitative results obtained from all AL methods on S3DIS and SemanticKITTI datasets. We compare our method \ours to the proposed baselines, random sets and also the 
random seed used in ReDAL's paper~\cite{wu2021redal} to produce results, noted ReDAL's seed in the table.

\vspace{-6pt}
\section{Ablation experiments}
\label{sec:ablation_exps}
 
\begin{figure*}[ht!]
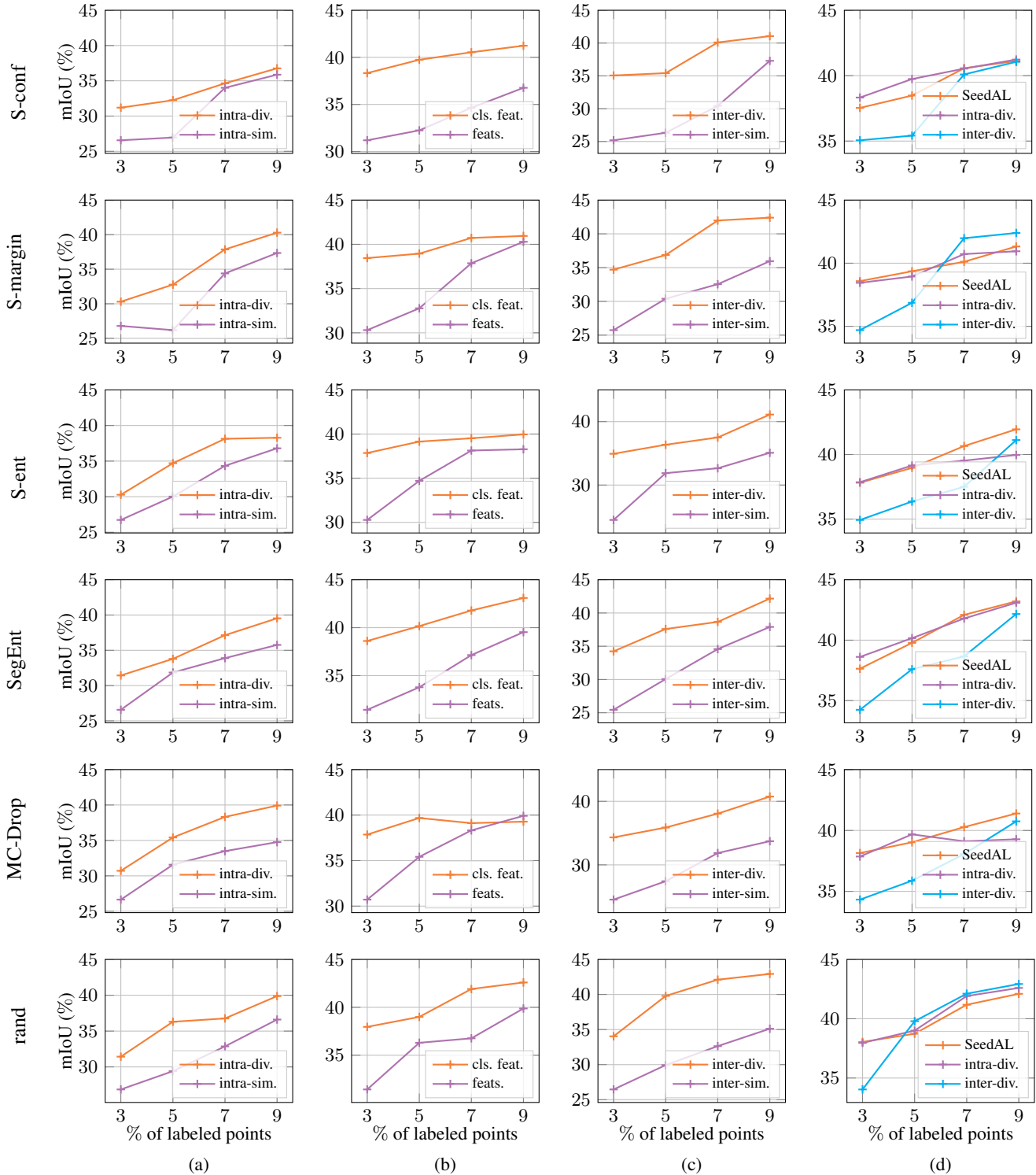

    \centering
    \include{figs/tikz/supp_ablation_study}
    \vspace*{-28pt}
    \caption{%
    \emph{Ablation study.} [Complement to Figure~7 in the paper] We evaluate here results obtained with different seeding strategies. (a)~Seeds made of scenes with high intra-diversity (intra-div.) or high intra-similarity (intra-sim.). (b)~Seeds selected with two different intra-diversity metrics:
    view features (feats.)
    or computed after clustering the view features (cls.\ feat.). (c)~Seeds made of scenes with high inter-diversity (inter-div.) or high inter-similarity (inter-sim.). (d)~Seeds selected with \ours, considering only inter-diversity (inter-div.) or intra-diversity (intra-div.).
    }
    \label{fig:ablation-rest}
\end{figure*}

\autoref{fig:ablation-rest} shows the remaining results of our ablation experiments obtained from  AL methods \random, \mcdropout, \softconf, \softmargin, \softentropy and \segent. These results corroborate what is presented in Figure~7 and Section~5.4 of the paper, namely that: (a)~intra-scene diversity is particularly relevant, compared intra-scene similarity; (b)~clustering features leads to better AL seeds than just exploiting intra-scene diversity; (c)~inter-scene diversity leads to better AL seeds than inter-scene similarity; (d)~the proposed combination of intra- and inter-scene diversity (i.e., \ours) generally performs on par or better than both intra- or inter-scene diversity, independently.

\begin{figure*}[ht]
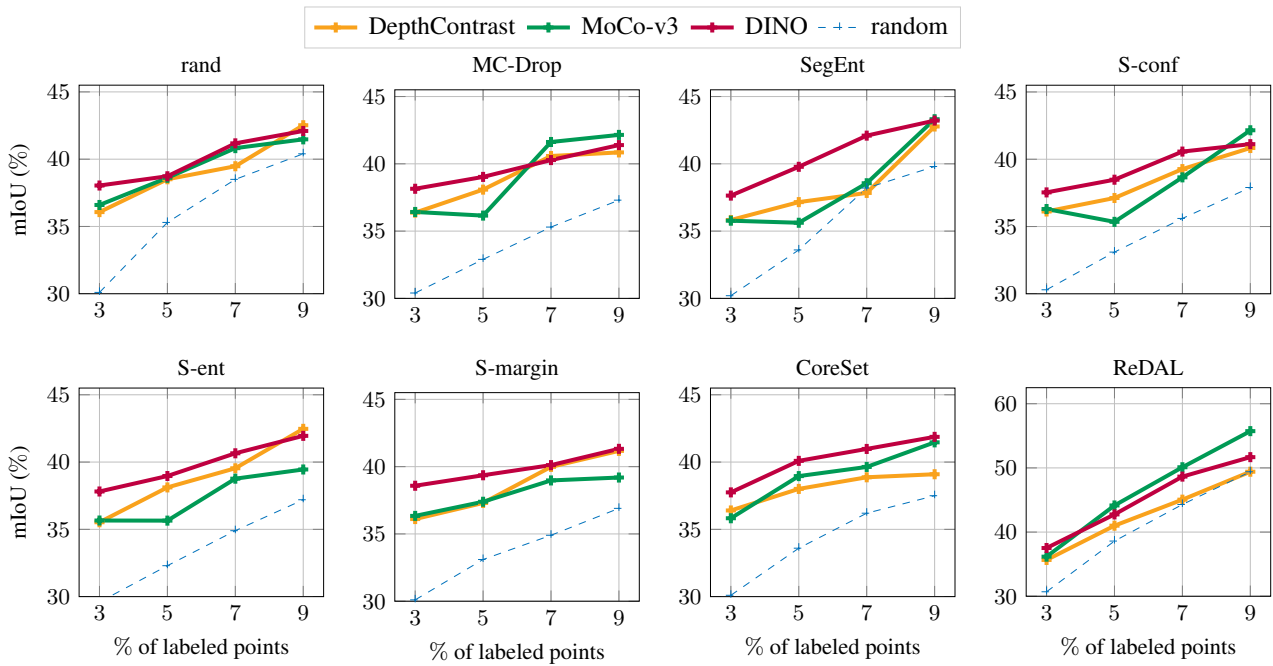

    \include{figs/tikz/s3dis-ss-feats}
    \vspace{-15pt}
    \caption{\nermin{SeedAL results on S3DIS using features from MoCo-v3 and DINO. Rand is an average over the random seeds.} 
    \vspace{-10pt}
    }
    \label{fig:other-ss-feats-s3dis}
\end{figure*}

\begin{figure*}[ht]
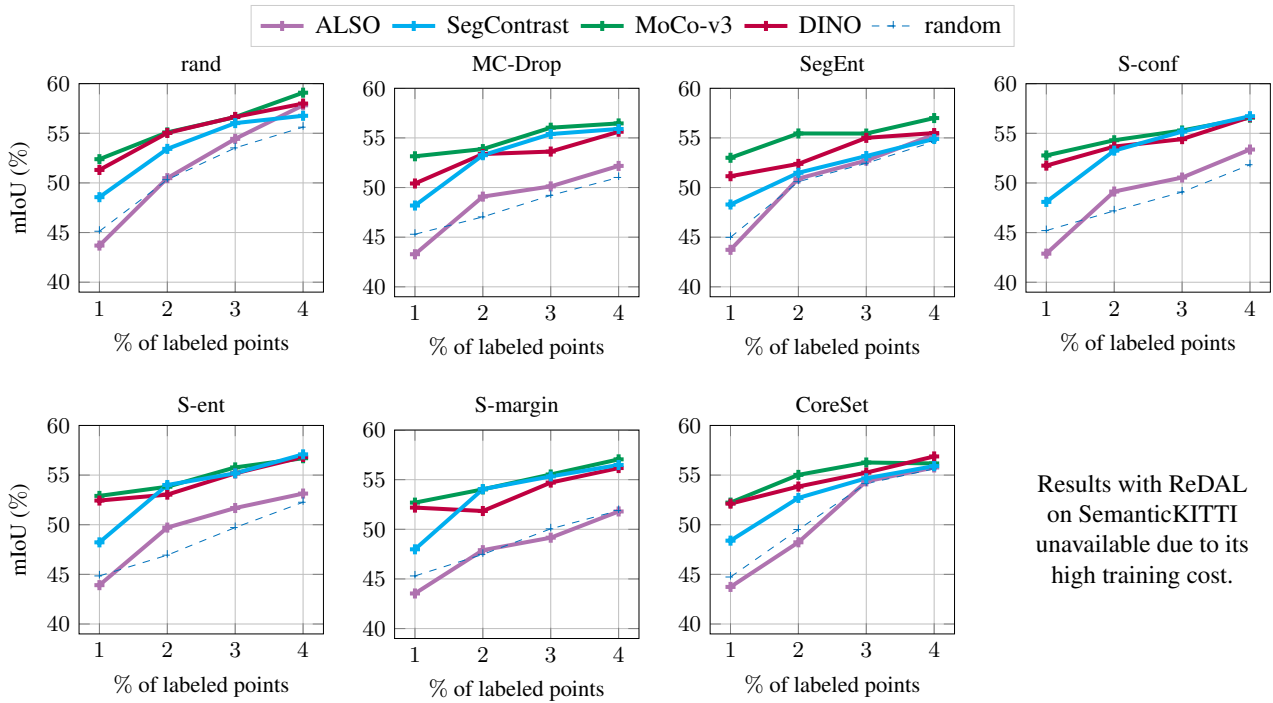

    \include{figs/tikz/kitti-ss-feats} 
    \vspace{-15pt}
    \caption{\nermin{SeedAL results on SemanticKITTI using features from DepthContrast, SegContrast, ALSO, MoCo-v3, DINO. Rand is an average over the random seeds.} 
    \vspace{-10pt}
    }
    \label{fig:other-ss-feats-kitti} 
\end{figure*}

\nermin{

As a complement to Figure~8 in the paper, \autoref{fig:other-ss-feats-s3dis} and \autoref{fig:other-ss-feats-kitti} present the results for all active learning methods with different 2D and 3D self-supervised features on S3DIS and SemanticKITTI, respectively. We do not provide results with ReDAL on SemanticKITTI due to its massive training cost.
}

\begin{table*}
    \centering
     \resizebox{.71\textwidth}{!}{
    
    \begin{tabular}{ll|rrrr|rrrr}
    \toprule
    && \multicolumn{4}{c|}{S3DIS} & \multicolumn{4}{c}{SemanticKITTI}
    \\
    AL & AL seeding & \multicolumn{4}{c|}{(\% of labeled points)} & \multicolumn{4}{c}{(\% of labeled points)}
    \\
    method & method & \multicolumn{1}{c}{3} & \multicolumn{1}{c}{5} & \multicolumn{1}{c}{7} & \multicolumn{1}{c|}{9} & \multicolumn{1}{c}{1} & \multicolumn{1}{c}{2} & \multicolumn{1}{c}{3} & \multicolumn{1}{c}{4} 
    \\
    \midrule
    \multirow{5}{*}{\relax{\random}}
    & random      & 30.1 & 35.3 & 38.5 & 40.4 & 46.1 & 50.8 & 53.9 & 55.9 \\[-1mm]
    & \qquad\footnotesize\color{MidnightBlue}{std dev} & \footnotesize\color{MidnightBlue}{5.5} & \footnotesize\color{MidnightBlue}{3.7} & \footnotesize\color{MidnightBlue}{2.9} & \footnotesize\color{MidnightBlue}{2.9} & \footnotesize\color{MidnightBlue}{3.5} & \footnotesize\color{MidnightBlue}{1.0} & \footnotesize\color{MidnightBlue}{1.5} & \footnotesize\color{MidnightBlue}{1.4} \\
    & KMcentroid & 33.5 & 36.9 & 39.8 & 40.9 & 41.0 & 48.9 & 53.6 & 54.7 \\
    & KMfurthest & 36.2 & 36.8 & 40.9 & \textbf{42.8} & 39.8 & 51.9 & 53.8 & 56.7\\
    & ReDAL's seed & 26.1 & 30.0 & 35.9 & 39.8 & 48.0 & 51.9 & 54.6 & 56.6 \\
    & \ours & \textbf{38.0} & \textbf{38.7} & \textbf{41.2} & 42.1 & \textbf{51.3} & \textbf{55.0} & \textbf{56.6} & \textbf{58.0} \\ 
    \midrule
    \multirow{5}{*}{\relax{\softconf} \cite{softmax}}    
    & random      & 30.3 & 33.1 & 35.6 & 37.9 & 46.1 & 48.2 & 49.9 & 52.6 \\[-1mm]
    & \qquad\footnotesize\color{MidnightBlue}{std dev} & \footnotesize\color{MidnightBlue}{5.8} & \footnotesize\color{MidnightBlue}{3.5} & \footnotesize\color{MidnightBlue}{3.0} & \footnotesize\color{MidnightBlue}{3.7} & \footnotesize\color{MidnightBlue}{3.7} & \footnotesize\color{MidnightBlue}{4.2} & \footnotesize\color{MidnightBlue}{3.8} & \footnotesize\color{MidnightBlue}{3.2} \\
    & KMcentroid & 33.4 & 33.5 & 36.7 & 38.2 & 41.4 & 48.6 & 50.1 & 53.4 \\
    & KMfurthest & 36.5 & 36.6 & 39.2 & 41.1 & 39.9 & 46.9 & 50.4 & 52.5\\
    & ReDAL's seed & 26.1 & 26.6 & 29.3 & 34.6 & 47.9 & 50.2 & 51.7 & 54.2 \\
    & \ours & \textbf{37.5} & \textbf{38.5} & \textbf{40.6} & \textbf{41.1} & \textbf{51.7} & \textbf{53.7} & \textbf{54.4} & \textbf{56.6} \\ 
    \midrule
    \multirow{5}{*}{\relax{\softmargin} \cite{softmax}}
    & random      & 30.1 & 33.1 & 34.9 & 36.9 & 45.6 & 48.3 & 50.1 & 51.6 \\[-1mm]
    & \qquad\footnotesize\color{MidnightBlue}{std dev} & \footnotesize\color{MidnightBlue}{5.6} & \footnotesize\color{MidnightBlue}{4.1} & \footnotesize\color{MidnightBlue}{4.0} & \footnotesize\color{MidnightBlue}{3.7} & \footnotesize\color{MidnightBlue}{3.3} & \footnotesize\color{MidnightBlue}{2.6} & \footnotesize\color{MidnightBlue}{2.4} & \footnotesize\color{MidnightBlue}{2.4} \\
    & KMcentroid & 33.6 & 36.1 & 36.5 & 39.1 & 40.5 & 46.6 & 49.1 & 50.4 \\
    & KMfurthest & 36.8 & 38.7 & 38.5 & 41.2 & 40.5 & 45.5 & 48.5 & 50.3 \\
    & ReDAL's seed & 26.1 & 28.3 & 35.5 & 39.9 & 46.2 & 50.0 & 50.3 & 50.9 \\
    & \ours & \textbf{38.6} & \textbf{39.4} & \textbf{40.1} & \textbf{41.3} & \textbf{52.2} & \textbf{51.8} & \textbf{54.7} & \textbf{56.2} \\ 
    \midrule
    \multirow{5}{*}{\relax{\softentropy} \cite{softmax}}
    & random      & 29.6 & 32.3 & 34.9 & 37.2 & 45.7 & 47.9 & 49.8 & 52.1 \\[-1mm]
    & \qquad\footnotesize\color{MidnightBlue}{std dev} & \footnotesize\color{MidnightBlue}{5.6} & \footnotesize\color{MidnightBlue}{5.2} & \footnotesize\color{MidnightBlue}{4.0} & \footnotesize\color{MidnightBlue}{3.1} & \footnotesize\color{MidnightBlue}{3.2} & \footnotesize\color{MidnightBlue}{4.2} & \footnotesize\color{MidnightBlue}{3.7} & \footnotesize\color{MidnightBlue}{3.3} \\
    & KMcentroid & 33.2 & 34.1 & 36.5 & 41.3 & 41.8 & 46.6 & 50.1 & 53.1 \\
    & KMfurthest & 35.8 & 35.7 & 38.0 & 41.1 & 39.8 & 47.6 & 49.5 & 51.2\\
    & ReDAL's seed & 27.4 & 29.9 & 32.9 & 38.3 & 47.4 & 50.0 & 50.1 & 51.8 \\
    & \ours & \textbf{37.8} & \textbf{39.0} & \textbf{40.7} & \textbf{41.9} & \textbf{52.4} & \textbf{53.0} & \textbf{55.2} & \textbf{56.8} \\ 
    \midrule
    \multirow{5}{*}{\relax{\coreset} \cite{coreset}}    
    & random  & 30.1 & 33.6 & 36.2 & 37.5 & 45.3 & 49.5 & 53.5 & 55.1 \\[-1mm]
    & \qquad\footnotesize\color{MidnightBlue}{std dev}  & \footnotesize\color{MidnightBlue}{5.5} & \footnotesize\color{MidnightBlue}{4.2} & \footnotesize\color{MidnightBlue}{3.4} & \footnotesize\color{MidnightBlue}{2} & \footnotesize\color{MidnightBlue}{3.7} & \footnotesize\color{MidnightBlue}{2.6} & \footnotesize\color{MidnightBlue}{1.4} & \footnotesize\color{MidnightBlue}{1.0} \\
    & KMcentroid & 33.5 & 37.2 & 36.6 & 39.2 & 40.6 & 46.6 & 52.5 & 54.5 \\
    & KMfurthest & 36.4 & 39.3 & \textbf{41.1} & 39.7 & 40.1 & 46.4 & 50.2 & 54.7\\
    & ReDAL's seed & 26.3 & 30.2 & 32.3 & 34.9 & 46.4 & 49.5 & 52.1 & 54.1 \\
    & \ours & \textbf{37.7} & \textbf{40.1} & 40.9 & \textbf{41.9} & \textbf{52.1} & \textbf{53.8} & \textbf{55.2} & \textbf{56.9} \\ 
    \midrule
    \multirow{5}{*}{\relax{\mcdropout} \cite{mcdropout}}
    & random  & 30.4 & 32.9 & 35.3 & 37.3 & 46.4 & 48.2 & 50.3 & 52.1 \\[-1mm]
    & \qquad\footnotesize\color{MidnightBlue}{std dev}  & \footnotesize\color{MidnightBlue}{5.5} & \footnotesize\color{MidnightBlue}{4.3} & \footnotesize\color{MidnightBlue}{3.0} & \footnotesize\color{MidnightBlue}{3.3} & \footnotesize\color{MidnightBlue}{3.3} & \footnotesize\color{MidnightBlue}{5.2} & \footnotesize\color{MidnightBlue}{4.5} & \footnotesize\color{MidnightBlue}{4.4} \\
    & KMcentroid & 33.5 & 33.6 & 37.5 & 39.7 & 40.9 & 48.3 & 50.7 & 52.3 \\
    & KMfurthest & 37.1 & 37.4 & 38.9 & \textbf{42.4} & 39.4 & 43.5 & 49.0 & 51.7\\
    & ReDAL's seed & 26.9 & 28.9 & 31.5 & 32.1 & 48.6 & 50.6 & 52.4 & 54.2 \\
    & \ours & \textbf{38.1} & \textbf{39.0} & \textbf{40.3} & 41.4 & \textbf{50.4} & \textbf{53.4} & \textbf{53.6} & \textbf{55.6} \\ 
    \midrule
    \multirow{5}{*}{\relax{\segent} \cite{seg_ent}}
    & random  & 30.2 & 33.6 & 38.2  & 39.8 & 45.4 & 50.6 & 52.4 & 54.2 \\[-1mm]
    & \qquad\footnotesize\color{MidnightBlue}{std dev}  & \footnotesize\color{MidnightBlue}{5.5} & \footnotesize\color{MidnightBlue}{2.5} & \footnotesize\color{MidnightBlue}{1.9}  & \footnotesize\color{MidnightBlue}{2.1} & \footnotesize\color{MidnightBlue}{3.8} & \footnotesize\color{MidnightBlue}{1.5} & \footnotesize\color{MidnightBlue}{0.4} & \footnotesize\color{MidnightBlue}{0.8} \\
    & KMcentroid & 33.7 & 34.3 & 35.9 & 38.0 & 41.2 & 50.4 & 52.1 & 53.9 \\
    & KMfurthest & 35.9 & 38.2 & 38.8 & 40.5 & 40.1 & 49.8 & 51.3 & 54.0 \\
    & \ours & \textbf{37.6} & \textbf{39.8} & \textbf{42.1} & \textbf{43.2} & \textbf{51.1} & \textbf{52.4} & \textbf{55.0} & \textbf{55.5} \\ 
    \midrule
    \multirow{5}{*}{\relax{\redal} \cite{wu2021redal}}    
    & random & 30.7 & 38.6 & 44.3 & 49.4 & 44.9 & 51.8 & 55.8 & 57.9 \\[-1mm]
    & \qquad\footnotesize\color{MidnightBlue}{std dev} & \footnotesize\color{MidnightBlue}{5.2} & \footnotesize\color{MidnightBlue}{1.0} & \footnotesize\color{MidnightBlue}{0.5} & \footnotesize\color{MidnightBlue}{0.7} & \footnotesize\color{MidnightBlue}{3.2} & \footnotesize\color{MidnightBlue}{1.8} & \footnotesize\color{MidnightBlue}{0.8} & \footnotesize\color{MidnightBlue}{0.6} \\
    & KMcentroid & 32.6 & 37.6 & 45.3 & 48.3 & 38.5 & \textbf{53.9} & {55.7} & 57.2 \\
    & KMfurthest & 35.1 & 41.5 & 47.5 & \textbf{51.7} & 38.3 & 48.6 & 57.0 & 58.6\\
    & ReDAL's seed & 24.9 & 37.5 & 43.8 & 45.5 & 46.1 & 53.8 & \textbf{56.7} & 58.4 \\
    & \ours & \textbf{37.5} & \textbf{42.8} & \textbf{48.6} & \textbf{51.7} & \textbf{50.5} & \textbf{53.9} & {55.8} & \textbf{58.9} \\ 
    \bottomrule
    \end{tabular}
    }
    \caption{Performance (\%\,mIoU) of the AL seeding methods on several AL methods for S3DIS and SemanticKITTI. Noted `random' is the average over three and six random seeds for S3DIS and SemanticKITTI respectively (we also report the standard deviation ``std dev'').
    ``ReDAL's seed'' is the random seed used in the experiments reported in ReDAL's paper \cite{wu2021redal}. We report the results for the ReDAL method obtained after our re-training.
    }
    \label{tab:quantitative}
\end{table*}

\clearpage


\end{document}